\documentclass[letterpaper, 10 pt, conference]{ieeeconf}  
\IEEEoverridecommandlockouts                              % This command is only needed if 
\usepackage{graphics,dblfloatfix} % for pdf, bitmapped graphics files
\usepackage{graphicx}
\usepackage{epsfig} % for postscript graphics files
\usepackage{tabularx}
\let\labelindent\relax
\usepackage{enumitem,kantlipsum}
\usepackage[tracking=true]{microtype}

\usepackage{soul} %to cross out the font - \st command
\usepackage{amssymb}  % assumes amsmath package installed
\usepackage{MnSymbol}

\usepackage{comment}
\usepackage[font=footnotesize]{caption}
\usepackage[font=footnotesize]{subcaption}

\usepackage[utf8]{inputenc}
\usepackage[export]{adjustbox}
\usepackage{wrapfig}
\usepackage{array}
\usepackage{multirow}
\usepackage{booktabs}
\usepackage{cite} %this responsible for the reference in the line [12]-[14] not [12],[13],[14] 

\usepackage{hyperref}

\usepackage[utf8]{inputenc}
\usepackage[export]{adjustbox}
\usepackage{wrapfig}
\usepackage{array}
\usepackage{multirow}
\usepackage{booktabs}
\usepackage{float}
\usepackage{cite} %this responsible for the reference in the line [12]-[14] not [12],[13],[14] 

\usepackage{dsfont}
\usepackage{url}
\usepackage{mathtools} %matrix alignment
\usepackage{xcolor}

\usepackage{lipsum}% http://ctan.org/pkg/lipsum

\usepackage{amsmath}
\usepackage{bm}
\newcommand{\bA}{\bm{A}}
\newcommand{\bM}{\bm{M}}
\newcommand{\bC}{\bm{C}}
\newcommand{\bg}{\bm{g}}
\newcommand{\bq}{\bm{q}}
\newcommand{\btau}{\bm{\tau}}

\newcommand{\bF}{\bm{F}}

\newcommand{\bJ}{\bm{J}}

\usepackage{xcolor}

\usepackage{dblfloatfix}

\newcommand{\norm}[1]{\lVert#1\rVert}

\graphicspath{ {images/} }
\title{\LARGE \bf
Hierarchical Whole-body Control of the cable-Suspended \\Aerial Manipulator endowed with Winch-based Actuation
}

\author{Yuri S. Sarkisov$^{1}$,  Andre Coelho$^{1,2,3}$, Maihara G. Santos$^{4}$, \\Min Jun Kim$^{5}$, Dzmitry Tsetserukou, Christian Ott$^{6}$, and Konstantin Kondak$^{1}$ % <-this % stops a space
	\thanks{$^{1}$The authors are with Institute of Robotics and Mechatronics, German Aerospace Center (DLR), Wessling, Germany.}
    \thanks{$^{2}$The author is with Robotics and Mechatronics Group, University of Twente, Enschede, The Netherlands.}
   \thanks{$^{3}$The author is with Dextrous Robotics Inc., Memphis, United States.}
    \thanks{$^{4}$The author is with Instituto Tecnológico de Aeronáutica, São José dos Campos, Brazil.}
    \thanks{$^{5}$The author is with Intelligent Robotic Systems Lab, Korea Advanced Institute of Science and Technology (KAIST), Daejeon, Korea.}
    \thanks{$^{6}$The author is with Automation and Control Institute, TU Wien, Vienna, Austria.}
	\thanks{
	{\tt\footnotesize e-mail: ysarkisov90@gmail.com%, minjun.kim@dlr.de, andre.coelho@dlr.de, d.tsetserukou@skoltech.ru, Christian.Ott@dlr.de, Konstantin.Kondak@dlr.de
	}}%
	}

\begin{document}
\maketitle
\thispagestyle{empty}
\pagestyle{empty}

%%%%%%%%%%%%%%%%%%%%%%%%%%%%%%%%%%%%%%%%%%%%%%%%%%%%%%%%%%%%%%%%%%%%%%%%%%%%%%%%
\begin{abstract}
During operation, aerial manipulation systems are affected by various disturbances. Among them is a gravitational torque caused by the weight of the robotic arm. Common propeller-based actuation is ineffective against such disturbances because of possible overheating and high power consumption. To overcome this issue, in this paper we propose a winch-based actuation for the crane-stationed cable-suspended aerial manipulator. Three winch-controlled suspension rigging cables produce a desired cable tension distribution to generate a wrench that reduces the effect of gravitational torque. In order to coordinate the robotic arm and the winch-based actuation, a model-based hierarchical whole-body controller is adapted. It resolves two tasks: keeping the robotic arm end-effector at the desired pose and shifting the system center of mass in the location with zero gravitational torque. 
%Since winch servos are position-controlled the admittance interface is accommodated to pass controller torque commands. 
The performance of the introduced actuation system as well as control strategy is validated through experimental studies.
%During operation, aerial manipulation systems are affected by various disturbances. Among them is a gravitational torque caused by the weight of the robotic arm. Common propeller-based actuation is ineffective against such disturbances because of possible overheating and high power consumption. To overcome this issue, in this paper we propose a winch-based actuation for the crane-stationed cable-suspended aerial manipulator. Three winch-controlled suspension rigging cables produce a varying cable tension distribution and, consequently, generate a wrench to reduce the effect of gravitational torque. In order to coordinate the robotic arm and the winch-based actuation, a model-based hierarchical whole-body controller is adapted. It resolves two tasks: keeping the robotic arm end-effector at the desired pose and shifting the platform center of mass in the location with zero gravitational torque. 
%Since winch servos are position-controlled the admittance interface is accommodated to pass controller torque commands. 
%The performance of the introduced actuation system as well as control strategy is validated through experimental studies.
%\red{Three control- lable rigging cables link the suspension point (the crane’s hook) with the platform and allow to change its translational pose. Such an actuation approach reduces the effect of disturbing gravitational torque caused by the robotic arm weigh distribu- tion on the aerial base.}
\end{abstract}

%%%%%%%%%%%%%%%%%%%%%%%%%%%%%%%%%%%%%%%%%%%%%%%%%%%%%%%%%%%%%%%%%%%%%%%%%%%%%%%%
\section{Introduction}
\label{sec:Intro}

In the past decade, the aerial manipulation ($AM$) field has attracted intense interest among researchers due to numerous prospective industrial applications \cite{ollero2021past, ruggiero2018aerial, khamseh2018aerial}. In general, the aerial manipulator can be defined as the aerial base with attached manipulation device, e.g., robotic arm. Due to the physics of the coupled system, various disturbances affect the aerial manipulator during operation. Rapidly evolving perturbations, e.g., caused by wind, can be efficiently handled by use of high bandwidth actuation, namely, propeller propulsion. However, quasistatic disturbances such as internal displacement of the system center of mass ($COM$) due to robotic arm weight cannot be compensated in the same manner because of risk of actuation overheating and high power consumption \cite{kudelina2020main, nath2017optimization}. Nevertheless, this type of perturbation significantly complicates % \red{for precision and robustness} 
the interaction tasks by affecting $AM$ performance and onboard sensor (IMU, camera) measurements. As a solution, researchers investigated the direct weight sliding at the aerial base%during the operation
, e.g., the motion of additional masses/battery \cite{haus2017, kim2018oscillation, alakhras2022design, pose2022adaptive}.

%\red{
%> The delineation of "dynamic" and "static" is unclear. Inertial forces (claimed to be dynamic) can be static and likewise external wrenches (claimed to be static) can be dynamic. This terminology or distinction should be improved.
%A2: rapidly evolving, inner motion vs disturbances !
%A3: due to the physics of the suspended aerial manipulator, static and dynamics disturbances occur during ma- nipulation that affect the system: pendulum-like cable oscillations due to external perturbation or the platform tilt because of the robotic arm weight. 
%A4: Dy- namic disturbances are expected to alter with high frequency. To this end, fully-actuated propeller-based propulsion, which can generate omnidirectional wrench, is utilized. 
%A5: Specify the COM displacement as the main disturbance due to arm motion or interaction with environment}
    
\begin{figure}[t]
	\centering
	\includegraphics[trim=0 0 0 0,clip, width=0.925\linewidth]{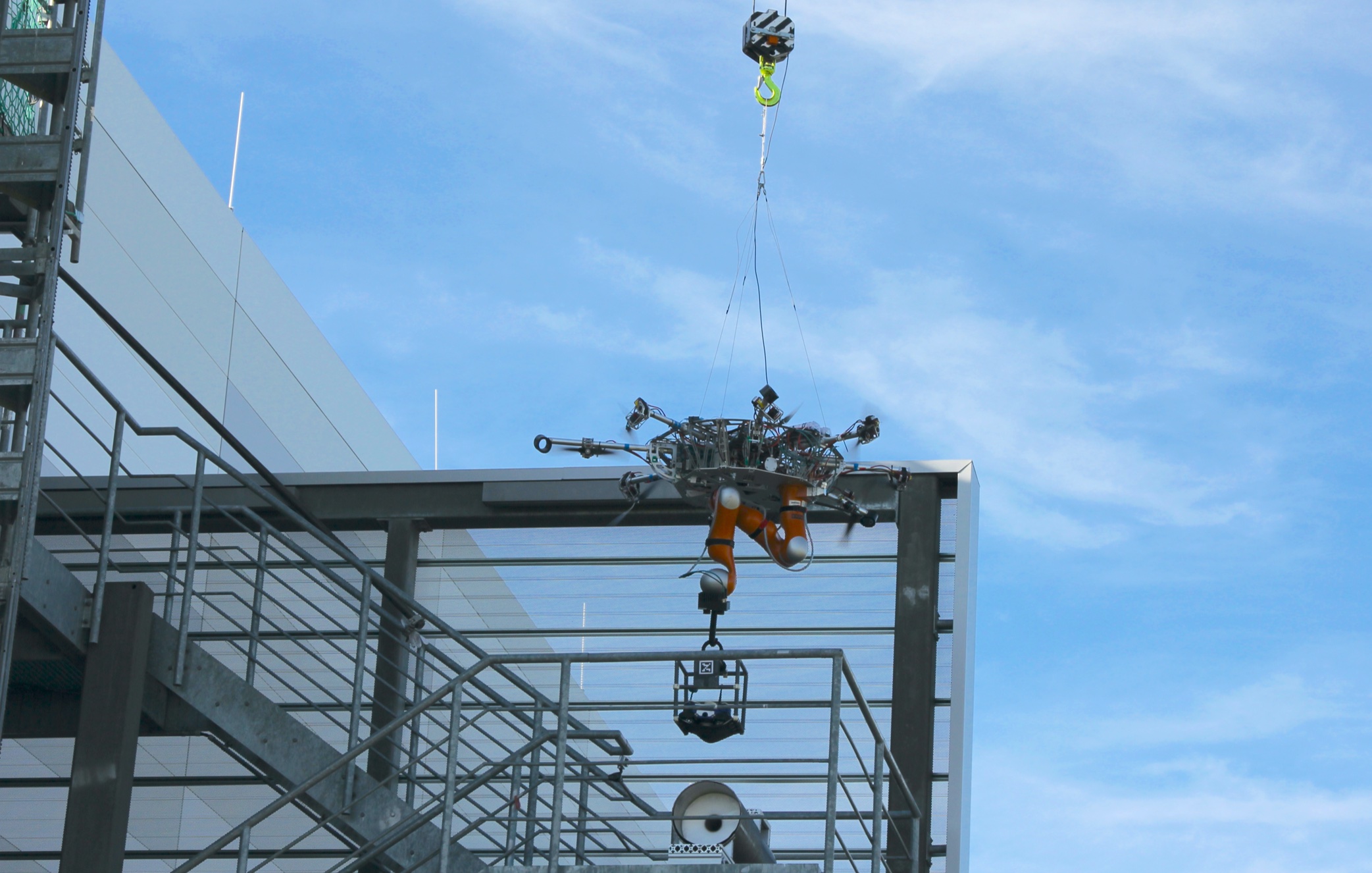}
	\caption{Cable-suspended aerial manipulator SAM.}
	\label{fig:main}
\end{figure}

In pursuit of increased performance and safety in the $AM$ field, cable-suspended aerial manipulators have been recently proposed\cite{miyazaki2019long, Yiit2021NovelOA, perozo2022optimal}. One example is the crane-stationed cable-Suspended Aerial Manipulator SAM developed in DLR \cite{sarkisov2019development}, see Fig. \ref{fig:main}. External crane suspension allows the platform to be compact and to utilize new approaches in order to deal with the % internal
displacement of $COM$.

\begin{figure*}[t]
	\captionsetup[subfigure]{labelformat=empty}
	\centering
	\subcaptionbox{(a) The SAM composition  \label{fig:real_sys} }{\includegraphics[height=4.35cm,keepaspectratio]{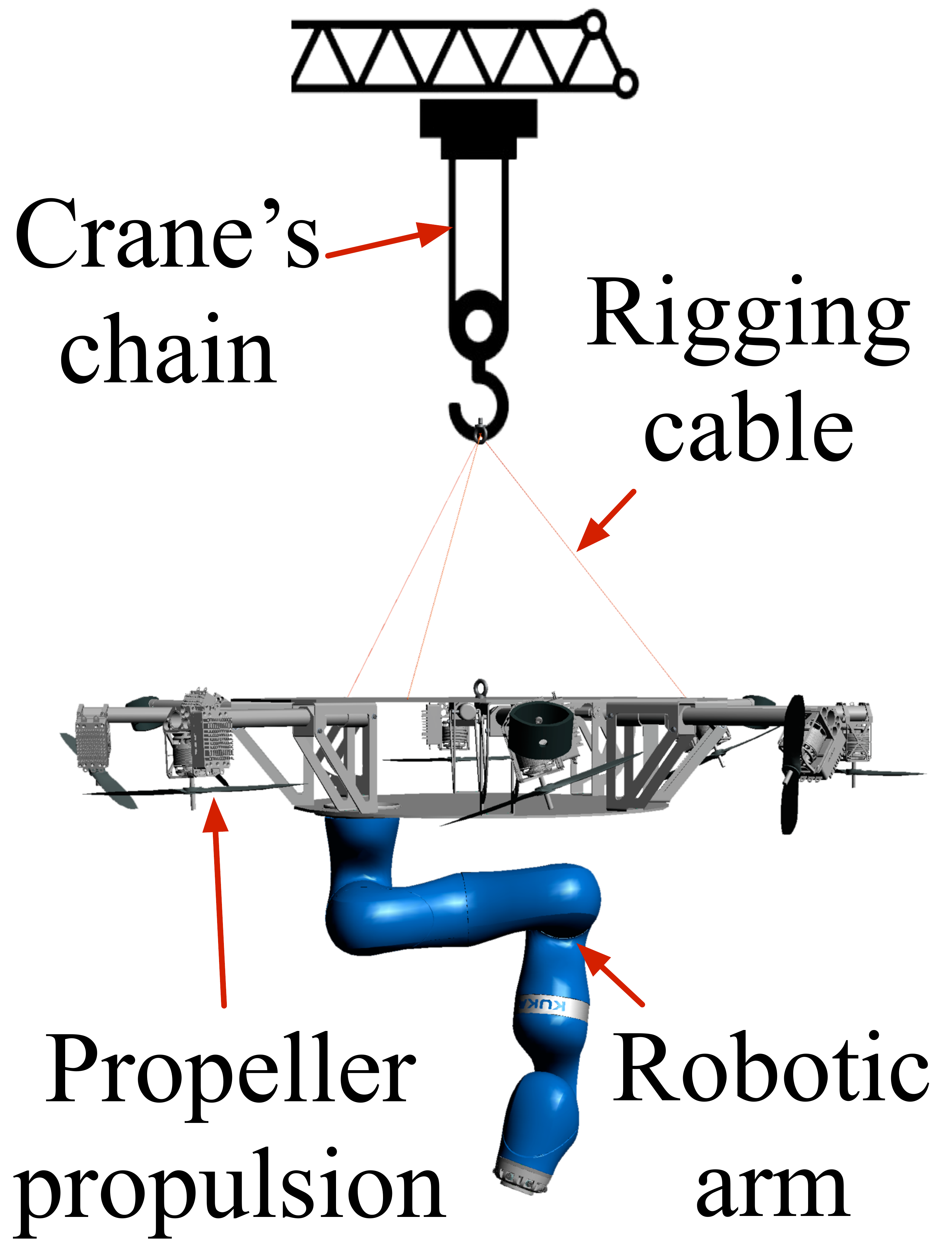}}
\subcaptionbox{(b) Three winches inside the platform  \label{fig:three_of_them} }{\includegraphics[height=4.35cm,keepaspectratio]{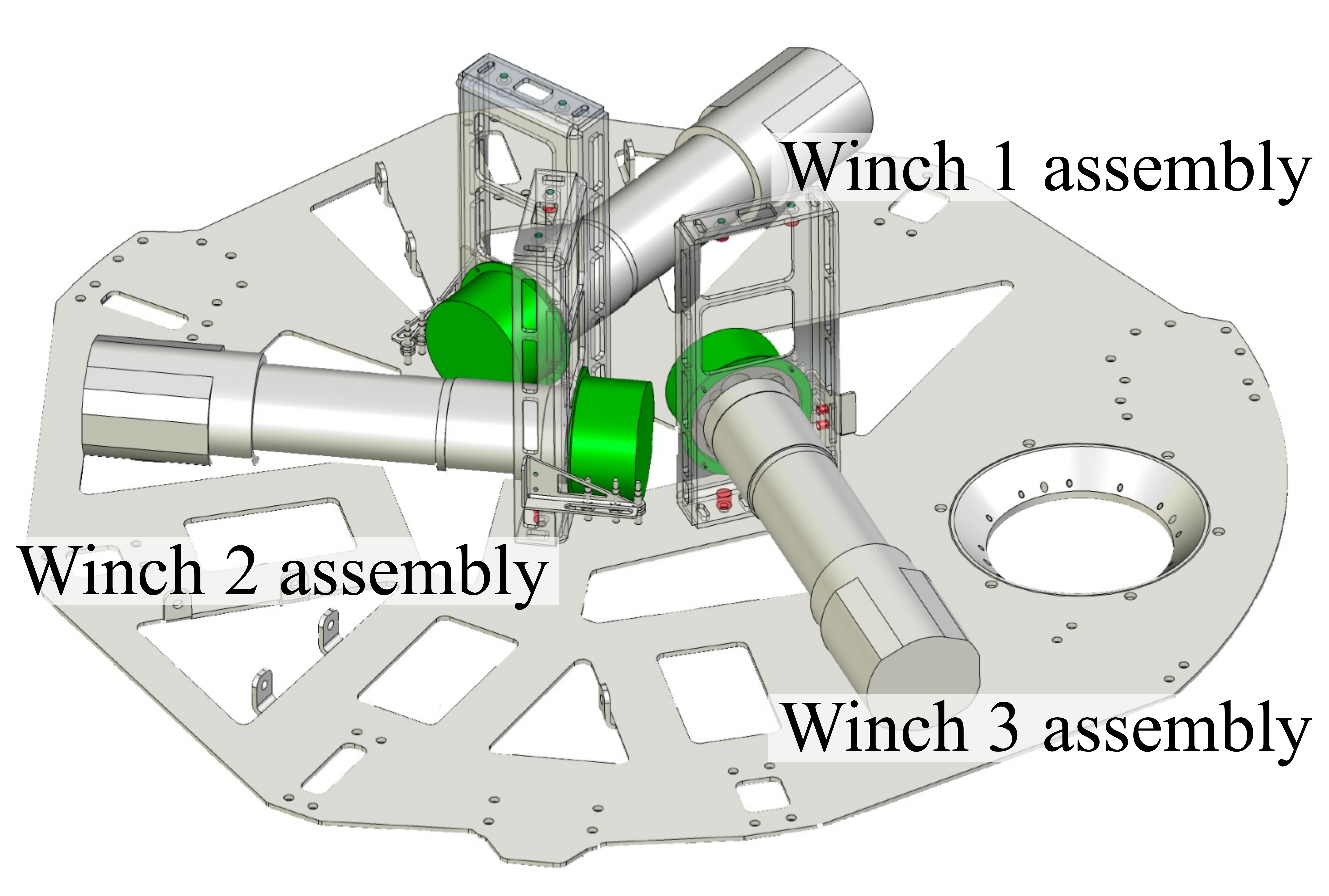}}
\subcaptionbox{(c) Main functional nodes  \label{fig:winches_op_diag} }{\includegraphics[height=4.35cm,keepaspectratio]{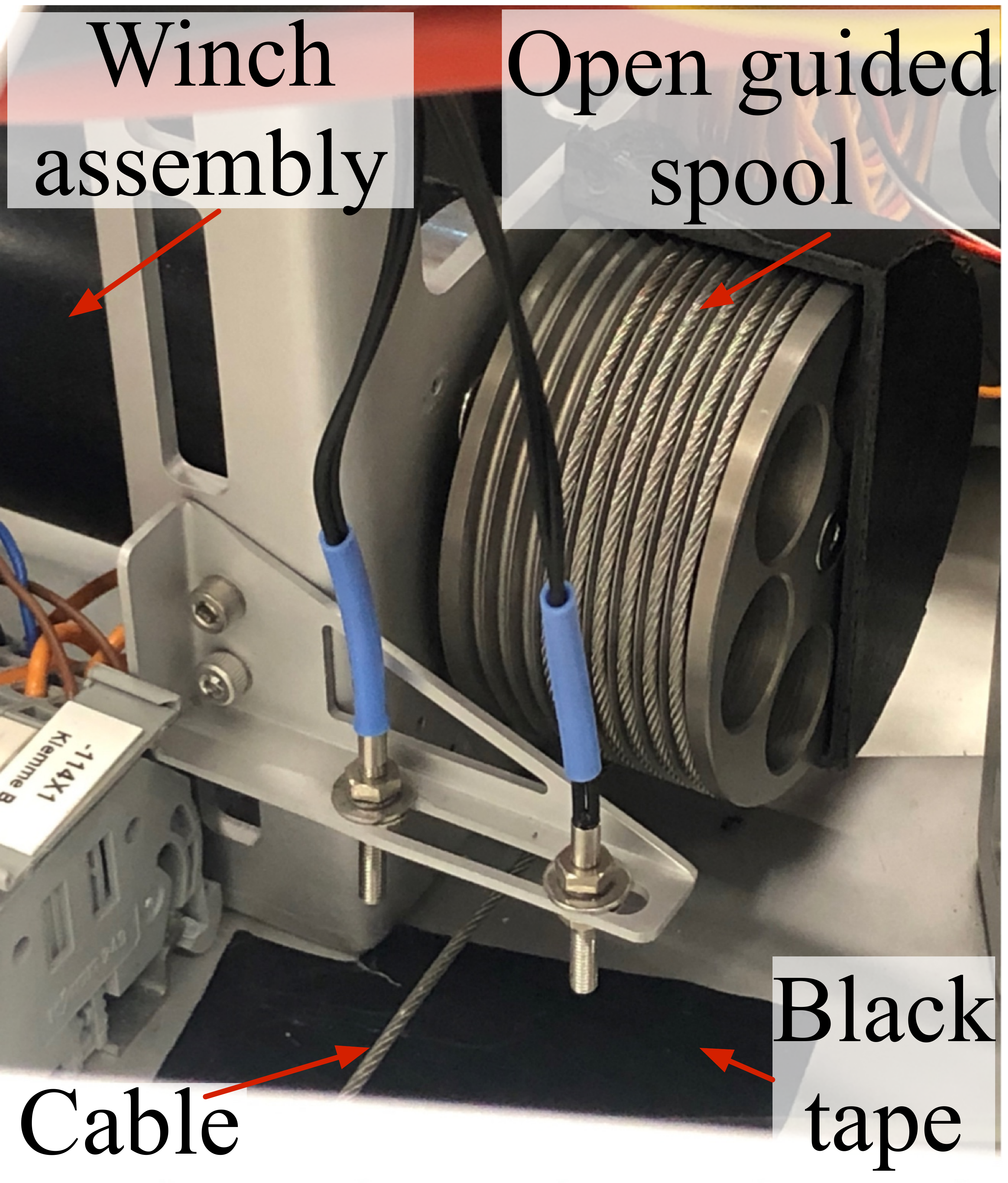}}
\subcaptionbox{(d) Cable detection process  \label{fig:winches_sensors} }{\includegraphics[height=4.35cm,keepaspectratio]{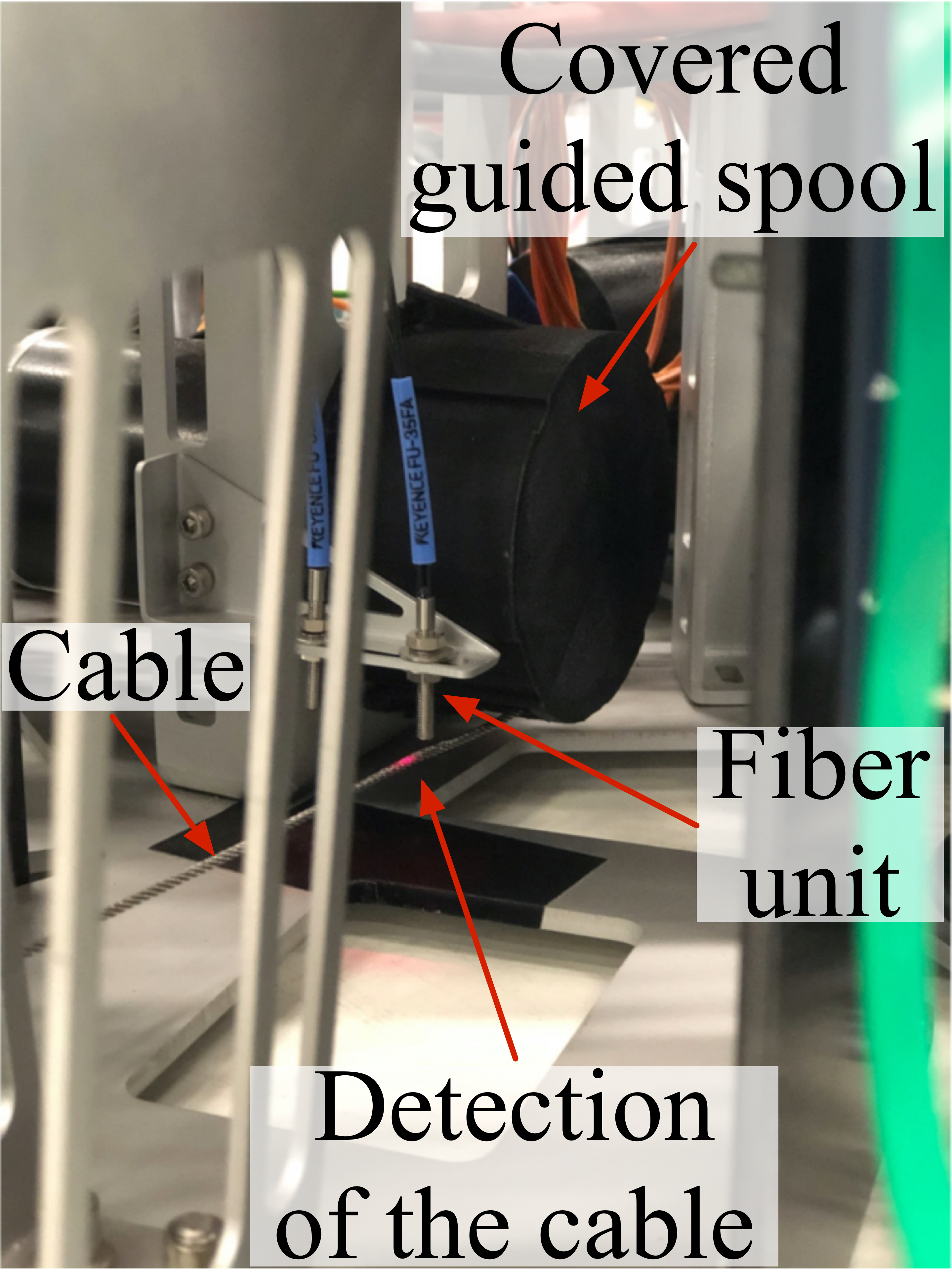}}
\caption{The winch-based actuation composition.}
	\label{fig:winch_figure}
\end{figure*}

In the scope of this paper, we introduce a novel winch-based actuation integrated to the SAM and inspired by the cable-driven robotics \cite{6907731, 1705581,begey2018dynamic}. %: design, modeling, control, and validation.
By regulating the length of three suspension rigging cables  connecting the SAM with the crane's hook, the winch-based actuation compensates for the gravitational torque generated by robotic arm weight. In order to design a controller for this  kinematically-redundant system, a number of challenges should be resolved. First of all, a closed-chain winch cabling in pair with attached robotic arm is modeled by Lagrangian constrained dynamics and mapped to the equal serial-chain coordinates describing the translational motion of the SAM base \cite{FREITAS2011, mcgrath2017lagrange, murray1989dynamic}. Secondly, the complex dynamics is simplified in order to find a reasonable trade-off between the system behavior description and equations that can be efficiently utilized in the controller. Further, a hierarchical whole-body controller with integrated admittance interface is adapted to the resulted kinematically redundant system extending our previous contribution \cite{coelho2021hierarchical} by adding capability of aerial base's translation motion in the null space.
% of the \red{integrating winches and adapting model for allowing translation motion of the base. Further Exploring null-space contro lcapabilities in adttion to the yaw control before for iproving filed of view now manipulation capabiltieis.}
%\red{First, the complex closed-chain winch cabling is modeled by Lagrangian constrained dynamics and mapped to the intuitive quasi-state [Blajer, 1997] defined by equal open-chain coordinates for further control formulation.}
% \red{dynamics reduction}
% \red{tradeoff between ... and ... was received }
%\red{control law that was first proposed here is extended…with correspnding modeling process and dynamic reduction  }
%\red{In order to coordinate robotic arm and winch dynamics, a Hierarchical impedance-based Whole-Body Controller is elaborated.}
The main aim of the controller is to perform two objectives with different priorities defined through the null-space: regulation of the robotic arm toward desired configuration and shifting system $COM$ to the location with zero torque due to gravity. Both tasks are regulated under impedance-based control. Benefits and performance investigation of the introduced winch-based actuation under the designed control strategy are studied through experiments.

\section{The SAM Platform}
\label{sec:SAM}
%In this section, we briefly describe the SAM platform and introduce its new actuation subsystem based on winches.

\subsection{General description}
\label{sec:SAMconc}
The SAM platform serves for the conduction of safe $AM$ in a complex industrial environment. During the operation, the system is suspended to the carrier, e.g., mobile crane, that brings the system to the operational point and compensates for total gravity. Attached to the bottom of the platform, a redundant 7 degrees of freedom ($DOF$) robotic arm KUKA LWR 4 performs the manipulation task while two actuation units, propeller propulsion and winches, keep the aerial base close to the motionless state, see Fig. \ref{fig:real_sys}. The former actuation contains small safe propellers that can generate the propeller-based omnidirectional wrench which is insufficient to counteract gravity but which can dampen oscillations caused by even small disturbances as was shown in \cite{sarkisov20}.

%due to low natural friction in a hook and suspension point, the SAM behaves as a double pendulum. After applying even small disturbances the platform keeps oscillating for a long time. To this end, in our previous work, we developed an oscillation damping controller that can dampen oscillations by exploiting  relying on single onboard IMU feedback. %Thus, agile propellers can be used to reduce dynamic disturbances.  
%\red{ with  total thrust from the multirotor . At the same time, the carrier at which the platform is suspended has high payloadcapability and compensates for the whole system gravity, so actuation efforts at the platform for stabilization should not be high. It allows to make the platform compact and safe. }
In order to further enhance the manipulation and stabilization performance of the system, in this work, we concentrate on the integration of the second actuation system, i.e., winches. It allows controlling the platform displacement relative to the winch suspension point via changing the length of the three rigging cables and, consequently, cable tension distribution. The winch actuation is only switched on % to compensate for the static deviations during the manipulation process 
after the SAM reaches the desired operational point and dampens oscillations by propeller propulsion. 

%Aerial manipulation with two actuations should allow avoiding overheating of the propeller propulsion and should reduce the amount of the consuming energy.
%It is worth mentioning that static disturbances affect the aerial base orientation, and as a result, cause the error in the robotic end-effector pose.

\subsection{Design of the winch-based actuation}
\label{sec:hardware}

The winch-based actuation contains three embedded winches each of which can control the length of the rigging cable that suspends the SAM platform to the crane's hook. %Intuitively, when the length of the cables is equal, the SAM can be seen as equilateral triangular pyramid edges of which represent the cables. The length of the cables can be controlled via embedded winches.
Each winch is installed inside the platform (Fig. \ref{fig:three_of_them}) and based on a motor assembly, that contains Maxon RE50 DC-motor, a planetary gearhead with a reduction ratio of 343/8, a 3 channel high precision encoder HEDL 9140 with 500 counts per turn, and a static brake AB 44. The regulation of each assembly is performed by controller EPOS4 compact 50/5 CAN with integrated motion controller based on the encoder feedback.

A guided spool is attached to the gearhead output shaft, at which the controlled steel cable is wounded, see Fig. \ref{fig:winches_op_diag}. The cable length varies from 0.5 to 1.3 meters. The sagging of the rigging cables within this range is unattainable even with horizontally stretched robotic arm. To keep length within this range, an optical fiber sensing system from Keyence is used. When the cable length exceeds the precalibrated limit range, the light beam of the fiber unit is interrupted as shown in Fig. \ref{fig:winches_sensors}, and the motor is paused.

% \begin{figure}[b]
%       \centering
%       \includegraphics[width=0.911\linewidth]{submission/images/WinchScheme_cut.pdf} 
%       \caption{\red{The communication diagram of the winch system.}}
%       \label{fig:winches_sche}
% \end{figure}
Three motor assemblies are integrated into a single communication Controller Area Network bus composed of can-high and can-low signals and terminated by a 120 $\Omega$ resistance on both sides. The communication is established via the higher layer CANopen protocol with a transmission rate of 1 Mbits/s. 
%Based on the CANopen addressing and communication scheme, each device has an object dictionary, which contains parameters that describe the behavior of the device, e.g., motor constants, controller gains. 
%This standard defines two communication objects. The first one is the Service Data Object used to configure the device. The second is the Process Data Object, which is exploited to control the motor and to read measurement data in real-time.

Onboard flight control computer ($FCC$) with deployed QNX Neutrino Real-time Operating System is utilized to establish the Leader/Followers communication between $FCC$ and EPOS4 controllers, a special library to manage the CANopen stack was developed. It allows operation of the winches with the desired rotational angle as an input.% The communication diagram is presented in Fig. \ref{fig:winches_sche}.  

\section{Modeling}
\label{sec: model}

In this section, we derive the SAM model that will be further used for control strategy formulation. We consider the case when the SAM is suspended to a fixed crane. To gain better intuition behind the complex system, first, we investigate a planar case, and further generalize it.

\subsection{Coordinate frames and main assumptions}
\label{sec:frames}
The SAM can be represented as a double pendulum suspended to the crane jib tip, $O$ (Fig. \ref{fig:points}). The length of the first link, ${OA}$, is equal to the length of the crane chain, $l_1$. We assume that the hook mass, $m_1$, is concentrated at point $A$. The second link consists of two objects: SAM platform and manipulator. Rigging cables, $AB$ and $AE$, are always under tension, so they can be represented as rigid links without sagging. Both tips of each cable, i.e., points $A$, $B$, $E$, are modeled as rotational joints.
The length of the rigging cables can be controlled and initially is equal to each other. The robotic arm is rigidly attached to the platform at point $D$. 

Let us introduce the following coordinate frames. 
The inertial frame $\mathcal{F}_w$ centered in the suspension point $O$ such that its vertical axis is opposite to downward gravity direction%\red{, and horizontal axis is perpendicular to it}
. 
%The hook frame $\mathcal{F}_h$ attached at point $A$ represents a rotation of the chain. %\red{The vertical axis of this frame is aligned with the chain, and the horizontal axis complements it to the right hand frame.} 
The platform frame $\mathcal{F}_p$ is aligned with the platform principal axes and placed at platform center, point $C$. A $COM$ frame $\mathcal{F}_{com}$ is located at the total $COM$ (platform and manipulator), at point $G$. %\red{The vertical axis of the frame is pointing toward suspension point $A$, and horizontal one is perpendicular to it.} %The $\mathcal{F}_{com}$ represents the rotation of the pendulum's second link relative to the first one, $q_2$.
%The last one is a tool frame $\mathcal{F}_t$ attached to the manipulator end-effector at point $F$. The orientation of the $\mathcal{F}_t$ is task dependent.

%50 0 10 0

\subsection{Dynamics formulation}
\label{sec:closed_chain}

\begin{figure}[t]
    % \begin{subfigure}[]{.22\linewidth}
    %   \centering
    %   % include second image
    %   \includegraphics[trim=0 0 0 0,clip, width=1\linewidth]{images/dp_model.pdf}  
    %   \caption{Double pendulum model}
    %   \label{fig:dp_model} 
    % \end{subfigure}
  \vspace{0.15cm}
    \begin{subfigure}[]{.44\linewidth}
        \centering
      % include second image
      \includegraphics[trim=0 0 0 0,clip, width=1\linewidth]{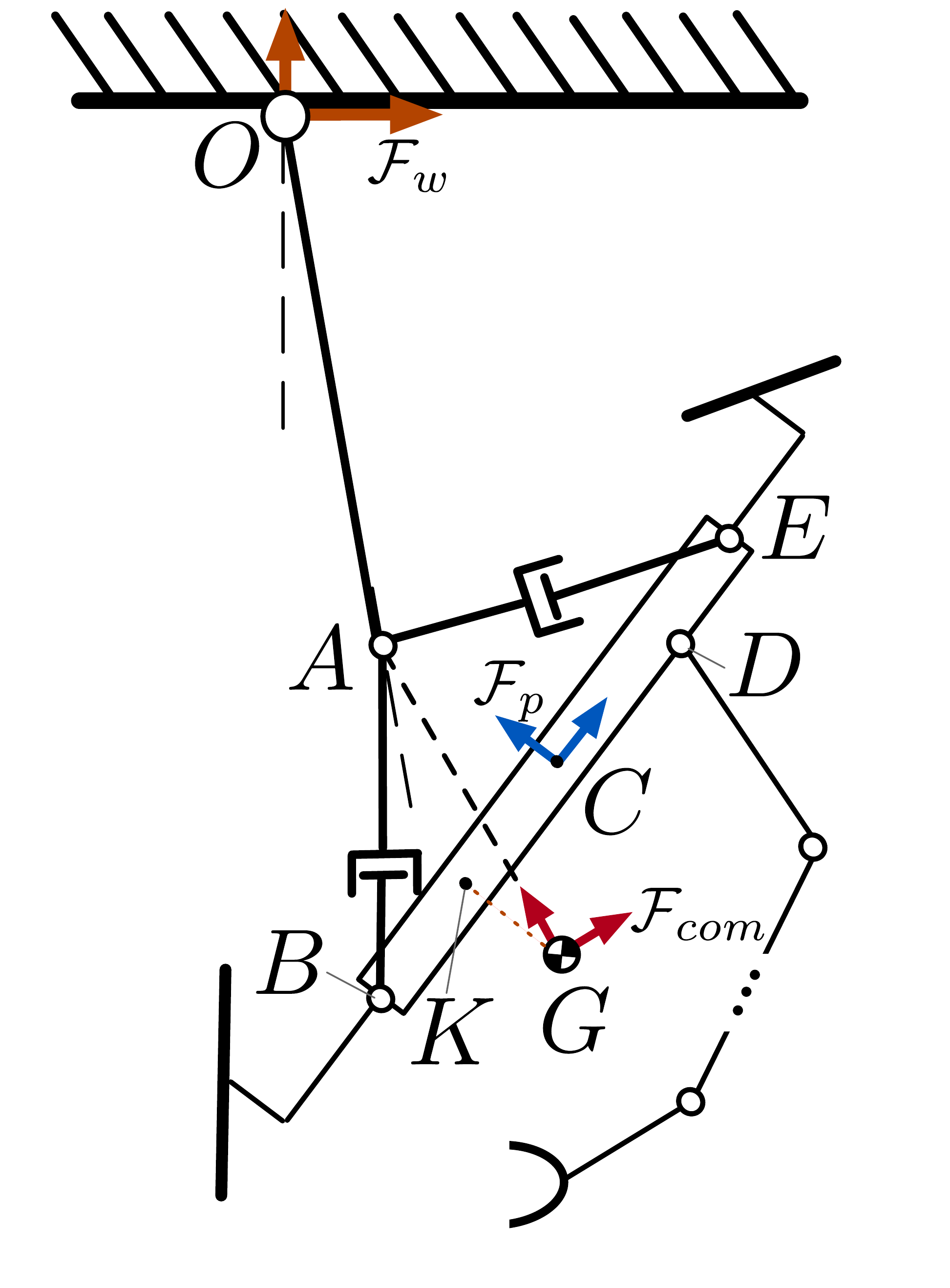} 
      \caption{Coordinate frames}
      \label{fig:points}
    \end{subfigure}
    ~ ~
    \begin{subfigure}[]{.44\linewidth}
      \centering
      % include second image
      \includegraphics[trim=0 0 0 0,clip, width=1\linewidth]{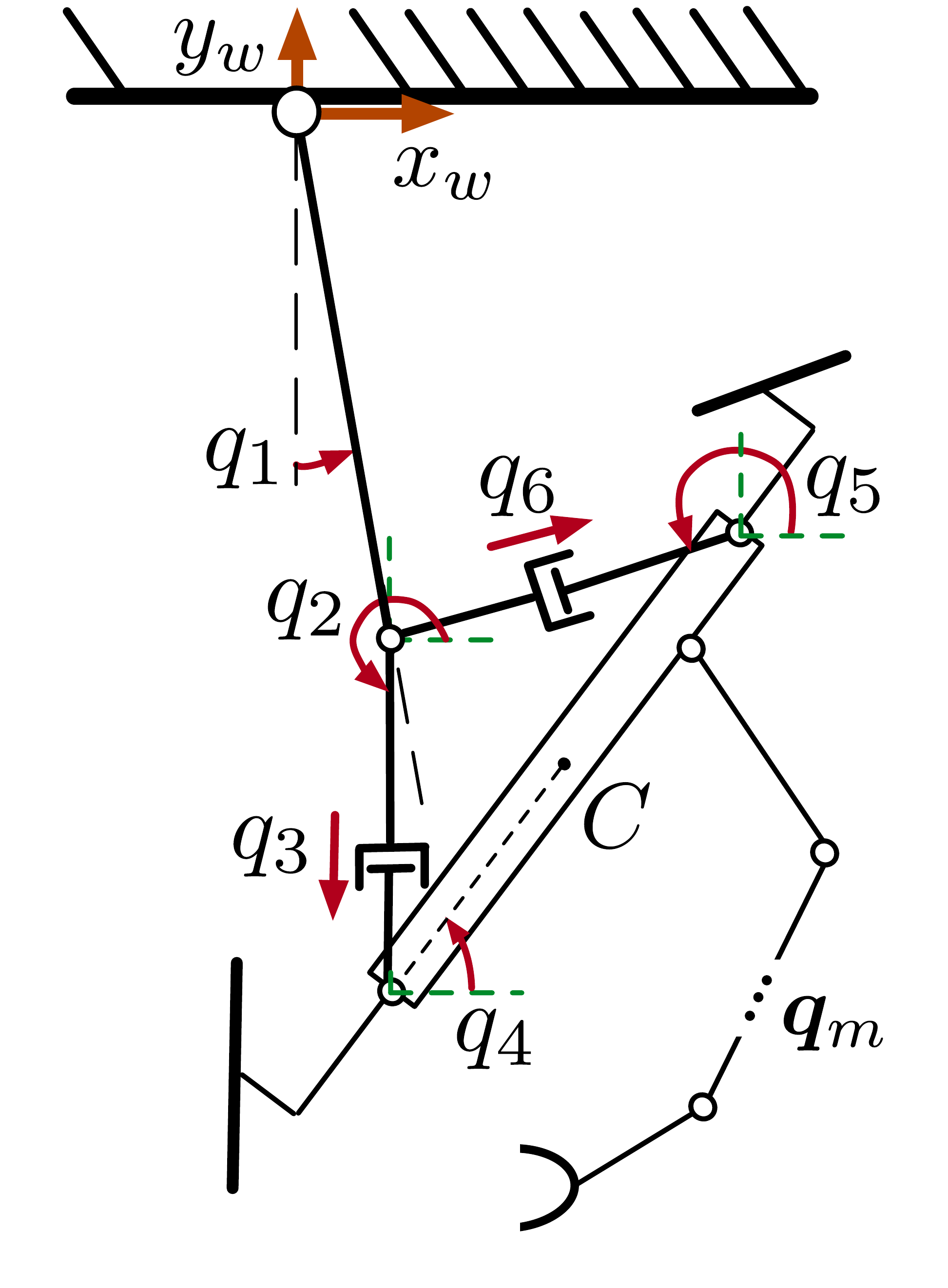} 
      \caption{Closed-chain coordinates}
      \label{fig:closed_chain}
    \end{subfigure}
    % ~ ~
    %     \begin{subfigure}[]{.22\linewidth}
    %   \centering
    %   % include second image
    %   \includegraphics[trim=0 0 0 0,clip, width=1\linewidth]{images/coordinates.pdf}
    %   \caption{Serial chain coordinates}
    %   \label{fig:serial_chain}
    % \end{subfigure}
    \caption{The SAM modeling.}
    \label{fig:model}
\end{figure}

The SAM planar model can be represented as a closed-chain system, see Fig. \ref{fig:closed_chain}. The corresponding state can be described by the vector of redundant generalized coordinates $\bq = [q_1 ,\ \bq_{winch}^{r} ,\ \bm{q}_m]^T \in \mathbb{R}^{6+m}$. Here, $\bq_{winch}^{r} = [q_2, q_3, q_4, q_5, q_6]^T$ is the vector of redundant winch-related joints including passive revolute joints, $q_2, q_4, q_5$, and active prismatic joints, $q_3$ and $q_6$, corresponding to the winch cable lengths, $\bq_m \in \mathbb{R}^{m}$ is the vector of the manipulator joint angles.
%It is worth noting that $q_2$ is not a part of the state, since it is a virtual quantity that might be computed from the system coordinates $\bq$ except for $q_1$. 

Since our model contains a closed chain $ABE$, the system dynamics is partially constrained. Applying Lagrangian formulation, the equation of motion can be written as\footnote{The term $\hat{\btau}_{ext}$ is omitted since we do not measure the external wrench applied to the system. In interaction tasks, a force torque sensor at the end-effector might be utilised for this purpose.}:% and assume that the designing controller is robust enough to cope with external perturbations.}:
\begin{align}
\begin{aligned}
    \label{eq:dynamics}
    \hat{\bM} (\bq) \ddot{\bq} + \hat{\bC} (\bq, \dot{\bq}) \dot{\bq} + \hat{\bg} (\bq) = \hat{\btau} + \bA^{T}\bm{\lambda}\\
    \bA(\bq)\dot{\bq} = 0, 
\end{aligned}
\end{align}
where $\hat{\bM}$ is the symmetric positive definite inertia matrix, $\hat{\bC}\dot{\bq}$ contains the centrifugal/Coriolis terms, $\hat{\bg}$ is the gravity vector, and $\hat{\btau}  \in \mathbb{R}^{6+m}$ is a vector of the joint torques. Let us denote by $n_w = 5$ a number of passive and active $DOF$s of the winch-related joints which cannot be controlled independently because of the $n_c = 2$ system constraints corresponding to the closed chain. Thus, there are only $n_p = n_w - n_c = 3$ $DOF$s along which platform can move using winch-based actuation.
%that  we  further omit  and  treat  as  a  source  of  disturbances  that  have  to  be dampened by controller. 
Then, $\bA \in \mathbb{R}^{n_c \times {6+m}} $ is a Pfaffian matrix, and $\bm{\lambda}\in\mathbb{R}^{n_c}$ are Lagrange multipliers that parametrize the interaction forces acting along the constraints.%, and $n_c = 2$ is the number of Pfaffian constraints. 

In order to impose holonomic constraints on the system dynamics, let us formulate a vector loop-closure equation that represents forward kinematics of the point $C = [x_c,\ y_c]^T$:
%\small
\begin{align}
\begin{aligned}
    \label{eq:constraint_vector_simplified}
    [x_c,\ y_c]^T = \overrightharpoon{OA} + \overrightharpoon{AB} + \overrightharpoon{BC}=
    \overrightharpoon{OA} + \overrightharpoon{AE} + \overrightharpoon{EC}.
\end{aligned}
\end{align}
%\normalsize

Projecting (\ref{eq:constraint_vector_simplified}) on $x_w$ and $y_w$ axes of the inertial frame and taking its time-derivative, we can derive two constraints in the Pfaffian form $\bA(\bq)\dot{\bq} = \bm{0}$. In order to resolve constraints and reduce the dimension of the state $\bq$, we apply a coordinate transformation.
To this end, let us define a vector of constraint-consistent independent generalized coordinates as follows: ${\bm{\delta}} = {[q_1,\ \bq_{winch},\ \bq_m}]^{T} = [q_1,\ {q}_3,\ {q}_4,\ {q}_6,\ \bq_m]^T  \in \mathbb{R}^{4+m}$. %and vector of dependent generalized coordinates as $\bm{z} = [q_3, q_6]^T\in \mathbb{R}^2$. 
Then, the relation between $\dot{\bm{\delta}}$ and $\dot{\bq}$ can be written as $\dot{\bq} = \bm{S}(\bq)\dot{\bm{\delta}}$ where $\bm{S}(\bq) \in \mathbb{R}^{(6+m) \times (4+m)}$ such that $\bm{S}^T(\bq) \bA^T(\bq) = 0$. 

As we mentioned above, there are $n_p = 3$ $DOF$s along which the platform can perform motion using winch system, i.e., $q_3, q_4, q_6$. In terms of practical application, we are interested in more intuitive motion directions,
e.g., platform horizontal $x_c$, vertical $y_c$, and rotational $q_4$ motions w.r.t. the inertial frame. To this end, by utilizing (\ref{eq:constraint_vector_simplified}), it is possible to express winch parameters $\bm{q}_{winch} = [q_3,\ q_4,\ q_6]^T$ as vector of the platform state $\bq_{p} = [x_c,\ q_4,\ y_c]^T \text{and } q_1$. Thus, we can define an equal state vector of feasible motions as $\bm{\eta} =  [q_1,\ \bq_{p},\ \bq_m]^T = [q_1, x_c, q_4, y_c, \bq_m]^T \in \mathbb{R}^{4+m}$, such that $\dot{\bm{\delta}} = \bm{B}(\bm{\delta})\dot{\bm{\eta}}$. 
Vector $\bm{\eta}$ corresponds to the equal serial-chain representation of the independent generalized coordinates defining the closed-chain dynamics. 

After coordinate transformation, the constrained dynamics (\ref{eq:dynamics}) can be formulated as unconstrained one in terms of $\bm{\eta}$ :
% \small
% \begin{align}
%     \label{eq:dynamics_fin}
%     \underbrace{{\bM}  (\bq)}_{\bm{B}^T \bm{S}^T \hat{\bM} \bm{S} \bm{B}}
%   \ddot{\bm{\eta}} 
%     +
%     \underbrace{ \bC (\bq,\dot{\bm{\eta}})}_{\bm{B}^T (\bm{S}^T \hat{\bM} \bm{S} \dot{\bm{B}}+ \atop \bm{S}^T (\hat{\bM} \dot{\bm{S}}+\hat{\bC}\bm{S})\bm{B})}
%     \dot{\bm{\eta}} 
%     + 
%      \underbrace{\bg (\bq)}_{\bm{B}^T \bm{S}^T \hat{\bg}}
%      =
%      \underbrace{{\btau}}_{\bm{B}^T \bm{S}^T \hat{\btau}}
%       + 
%      \underbrace{{\btau}_{ext}}_{\bm{B}^T \bm{S}^T {\hat{\btau}}_{ext}}.
% \end{align}
% \normalsize
\begin{align}
    \label{eq:dynamics_fin}
    {\bM}  (\bq)
   \ddot{\bm{\eta}} 
    +
    \bC (\bq,\dot{\bm{\eta}})
    \dot{\bm{\eta}} 
    + 
    \bg (\bq)
     =
     {\btau}.
    %  \underbrace{\bg (\bq)}_{\bm{B}^T \bm{S}^T \hat{\bg}}
    %  =
    %  \underbrace{{\btau}}_{\bm{B}^T \bm{S}^T \hat{\btau}}.
\end{align}
Here, ${\bM}(\bq) = \bm{B}^T \bm{S}^T \hat{\bM} \bm{S} \bm{B}$, $\bC (\bq,\dot{\bm{\eta}}) = \bm{B}^T (\bm{S}^T \hat{\bM} \bm{S} \dot{\bm{B}}+ \bm{S}^T (\hat{\bM} \dot{\bm{S}}+\hat{\bC}\bm{S})\bm{B})$, $\bg (\bq) = \bm{B}^T \bm{S}^T \hat{\bg}$, and ${\btau} = \bm{B}^T \bm{S}^T \hat{\btau}$. Vector of generalized torques applied to the serial dynamics (\ref{eq:dynamics_fin}) can be defined as $\btau = [\tau_{1},\  {\btau}_{p},\ \btau_{m}]^T \in \mathbb{R}^{4+m}$, where $\tau_{1}$ is a torque applied to the pendulum first joint, ${\btau}_{p}=[\tau_{x_c},\ \tau_{q_4},\ \tau_{y_c}]^T$ is a vector of control inputs applied to the joints corresponding to the platform motion, and $\btau_m$ are torques applied to the joints of the robotic arm.
%Since in the winch-based actuation we have only two actuated joints, $q_4$ and $q_7$, we should select two out of three quasi joints that we aim to control. In our case, $x_c$ and $y_c$ are of main interest. Thus, the vector of actuated torques, $\btau_a =[\tau_{q_1},\ \tau_{x_c},\ \tau_{y_c},\ \btau_m]^T\in \mathbb{R}^{3+m}$ such that $\btau = \bm{O}^T\btau_a$. Here, $\bm{O} \in \mathbb{R}^{(3+m) \times (4+m)}$ maps actuated quasi torques $\btau_a$ to the generalized torques $\btau$. 
For more details on $\bm{S}$, $\bm{A}$, $\bm{B}$ %, and $\bm{O}$ 
the reader is referred to \cite{ibrahim2007kinematic, my2019kinematic, zambella2019dynamic}.

% After coordinate transformation, unconstrained dynamics in terms of $\bm{\delta}$, can be formulated as:
% \small
% \begin{align}
%     \label{eq:dynamics_simplified}
%     \underbrace{\bar\bM (\bq)}_{\bm{S}^T \hat{\bM} \bm{S}}
%     \ddot{\bm{\delta}} + 
%     \underbrace{\bar\bC (\bq,\dot{\bm{\delta}})}_{\bm{S}^T (\hat{\bM} \dot{\bm{S}}+\hat{\bC}\bm{S})}
%     \dot{\bm{\delta}} + 
%     \underbrace{\bar\bg (\bq)}_{\bm{S}^T \hat{\bg}}
%     = 
%     \underbrace{\bar{\btau}}_{\bm{S}^T \hat{\btau}}
%     +
%     \underbrace{\bar{\btau}_{ext}}_{\bm{S}^T {\hat{\btau}}_{ext}}.
% \end{align}
% \normalsize

\begin{figure*}[t]
      \centering
      % include second image
      \includegraphics[width=0.9\linewidth]{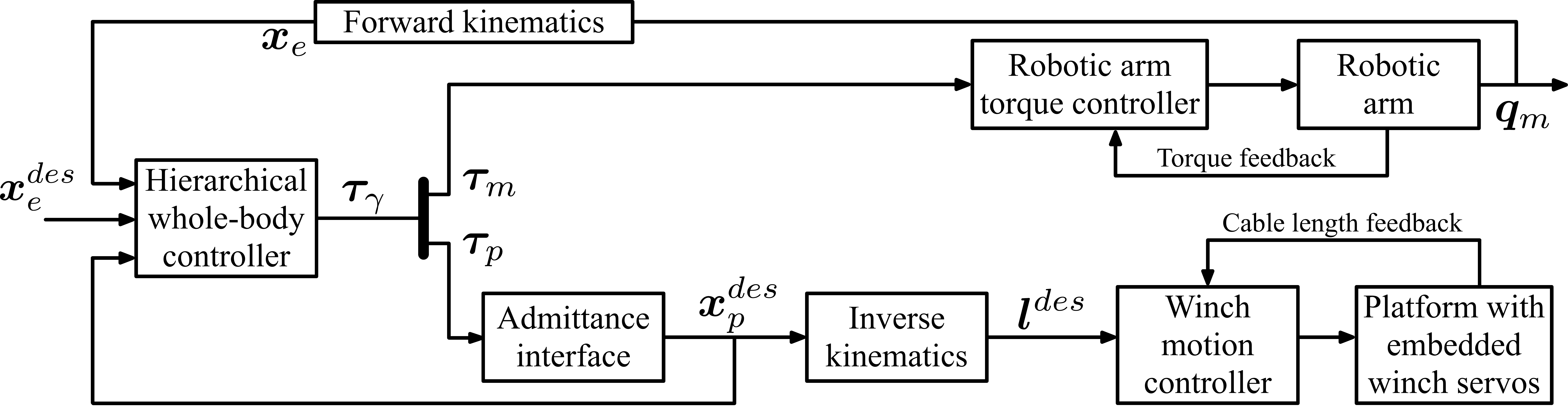}  
      \caption{Control diagram.}
      \label{fig:winch_control_diagram}
\end{figure*}

\subsection{Reduced model for control design}
\label{sec:3d_general}

Since winches cannot affect the motion of the platform around the yaw, a 3D case might be represented as two decoupled planar dynamics similar to the considered above. Therefore, the first joint of the double pendulum can be modeled by spherical joint, tips of the rigging cables can be modeled as universal joints with 2 $DOF$s at each fixation point, and three closed-chain loops provide 9 constraint equations. The platform can be moved by winch-based actuation in the space along 3 translational and 2 rotational $DOF$s corresponding to the roll and pitch of the platform. %Then, a quasi-state vector of feasible motions can be rewritten as: $\bm{\xi} = [\bm{q}_{s,pend_1},\ \bm{x}_{p},\ \phi ,\ \theta,\ \bm{q_m}]^T \in \mathbb{R}^{15}$, where m = 7 for redundant robotic arm.
%, and resulted dynamics in terms of the quasi state vector in 3D space can be formulated as:
% \small
% \begin{align}
% 	\label{eq:dynamics_simplified_final}
% 	\bM_s(\bm{\xi}) \ddot{\bm{\xi}} + \bC_s(\bm{\xi}, \dot{\bm{\xi}}) \dot{\bm{\xi}} + \bg_s (\bm{\xi})
% 	=
% \bm{O}^T_s \btau_{s,a}
% 	+ 
% {\btau}_{s,ext}.
% \end{align}
% \normalsize
% Here, $\btau_{s,a} = [\btau_{s,pend_1} ,\
% 			\tau_{x_{c}},\
% 			\tau_{y_{c}},\
% 			\tau_{z_{c}},\
% 			\btau_{m}]^T$.
%All values with subscript '$s$' correspond to the values of dynamics (\ref{eq:dynamics_fin}) generalized for the 3D case. %For further convenience, let us also define a vector of the spherical joint $\bm{q}_{s,pend_2}\in \mathbb{R}^{3}$ corresponding to the motion of the second link of the double pendulum, similarly to the $q_2$ for the planar case.

In the case, when oscillation damping controller is applied to the system \cite{sarkisov20}% (\ref{eq:dynamics_simplified_final})
, the dynamics of the pendulum joints ($q_1$ for planar case) can be neglected%, i.e., $\dot{\bm{q}}_{s,pend_1} = 0$ and $\dot{\bm{q}}_{s,pend_2} = 0$
, and corresponding pendulum joint angles will be defined over time only by gravitational torque depending on whether or not the system $COM$ is under the suspension point. Moreover, the winch-based actuation contains only three actuated joints, so we should select three out of five $DOF$s that we aim to control. In our case, translational motion is of main interest. Indeed, we can neglect by platform rotational dynamics since we can generate only such cable lengths that nullify platform tilting angles while keeping desired platform displacements as will be shown in \ref{subsec:ik}. Thus, the winch-based actuation allows translational motions of the platform w.r.t. the suspension point in order to counterbalance $COM$ displacement and, consequently, to nullify pendulum joints. To this end, we define a reduced feasible configuration $\bm{\gamma}$ which contains only joints responsible for the system $COM$ location and robotic arm end-effector pose as $\bm{\gamma} = [ \bm{x}_{p},\ \bm{q_m}]^T = [x_c, y_c, z_c, \bm{q_m}]^T$, here m = 7 for redundant robotic arm. 

Based on the aforementioned assumptions, the system dynamics can be formulated in terms of the $DOF$s which will be accessed by the controller as follows:
\begin{align}
	\label{eq:dynamics_simplified_final3}
	\bM_r(\bm{\gamma}) \ddot{\bm{\gamma}} + \bC_r(\bm{\gamma}, \dot{\bm{\gamma}}) \dot{\bm{\gamma}} + \bg_r (\bm{\gamma})
	=
\btau_\gamma.
\end{align}
Here, $\btau_\gamma = [\btau_p,\ \btau_m]^T$, and all elements with subscript ``$r$" correspond to the reduced dynamics.

\section{Control Approach}
\label{sec:ctrl}

In order to regulate the defined state $\bm{\gamma}$, in this chapter we adapt the hierarchical impedance-based whole-body controller \cite{dietrich2019hierarchical, ott15, dietrich2016whole},  see Fig. \ref{fig:winch_control_diagram}. As input, we set the desired pose of the end-effector, $\bm{x}_e^{des}$, while desired horizontal terms of the system $COM$, $\bm{x}_{com}^{des}$, are set to zero during the whole operation. The whole-body impedance-based controller produces the desired generalized torques for the robot joints, $\btau_m$, and joints of the platform, $\btau_p$. The robotic arm is directly torque-controlled and provides the end-effector position, $\bm{x}_e$, as an output. The platform with integrated winches on another hand is kinematically controlled. To this end, we first transform the desired generalized torques $\btau_p$ to the required motion along platform joints, $\bm{x}_{p}^{des}$, by virtue of admittance interface, and further we use inverse kinematics ($IK$) to map it to the desired cable lengths. In parallel, $\bm{x}_{p}^{des}$ is used in the controller in order to update the location of the resulted system $COM$. In the following subsections, all aforementioned sub-blocks will be considered in detail.

\subsection{Hierarchical decoupling and task space definition}
\label{sec:cg}

Let us define the following control tasks for our system:

\begin{enumerate}[wide, labelwidth=!, labelindent=0pt]
    \vspace{0.05cm}
    \item The main task for aerial manipulator is to change end-effector Cartesian pose%(three coordinates and three Euler angles)
    , $\bm{x}_{e} \in \mathbb{R}^{6}$, expressed in the inertial frame toward desired value, $\bm{x}^{des}_{e}$. Let us define configuration of the end-effector in terms of $\bm{\gamma}$ as $\bm{x}_{e} = \bm{f}_{1}(\bm\gamma)$, then our control goal is to provide: $\bm{x}_{e}-\bm{x}^{des}_{e}=0$.
    \vspace{0.15cm}
    \item The second task is to neutralize gravitational torque by keeping system $COM$, $\bm{x}_{com} \in \mathbb{R}^{3}$, under suspension point.
	Let us define the model $COM$ location in inertial space as $\bm{x}_{com} = \bm{f}_{2}(\bm{\gamma})$, then our control goal is to keep horizontal components of this vector, i.e., $x_{com}$ and $y_{com}$ as zeros. % This control goal ensures $\bq_{pend} = 0$.
    \vspace{0.15cm}
	\item The last task is injection of additional damping to any robotic arm joint (we chose the 3rd one, i.e., elbow). This task is introduced for further convenience.
\end{enumerate}
%The second task indirectly ensures that unmeasured pendulum joints are close to zero when oscillation damping controller is applied. Only in this case, we can state that the first task is fulfilled in the \emph{inertial frame} since utilized model is based only on the reduced state $\bm\gamma$.

It is worth mentioning, that in real mission, the robotic arm is under remote operator control \cite{lee20}, so the second task should have less priority \cite{dietrich2013multi}, and it should not interrupt the first (main) one. The third task has the least priority.
% Moreover, both control tasks should be structurally feasible at the same time.

Following the task space definition, we can define task velocities:
\begin{align}
	\label{eq:task_space}
	\begin{aligned}
		\bm{\dot{x}}_{e} = \bm{J}_1\dot{\bm\gamma}, \hspace{1cm}
		\bm{\dot{x}}_{com} = \bm{J}_2\dot{\bm\gamma}, \hspace{1cm}
		\dot{q}_{m,3} = \bm{J}_3 \dot{\bm\gamma},
	\end{aligned}
\end{align}
here $\bm{\dot{x}}_{e} \in \mathbb{R}^{6}$ is the body velocity of the end-effector, $\bm{\dot{x}}_{com} \in \mathbb{R}^{3}$ is the vector of translational velocities of model $COM$, $\dot{q}_{m,3} \in \mathbb{R}$ is the velocity of the robotic arm elbow joint. Besides,
$\bm{J}_i = \frac{\partial\bm{f}_i(\bm\gamma)}{\partial\bm\gamma}$ for $i \in \{1,\ 2,\ 3 \}$, i.e., $\bm{J}_{1} \in \mathbb{R}^{6 \times 10}$, $\bm{J}_{2} \in \mathbb{R}^{3 \times 10}$, and $\bm{J}_{3} \in \mathbb{R}^{1 \times 10}$ are corresponding Jacobian matrices that have a full row rank.

In order to avoid interference between tasks, we need to decouple the task velocities at the acceleration level. To this end, dynamically consistent null-space projectors $\bm{N}(\bm\gamma)$ \cite{khatib1987unified, dietrich2019hierarchical} have been used to dynamically decouple original Jacobian matrices in (\ref{eq:task_space}) as follows:
$
	\bm{\bar{J}}_i(\bm\gamma) = \bm{J}_i(\bm\gamma) \bm{N}_i(\bm\gamma)^T.
$
% \begin{align}
% 	\bm{\bar{J}}_i(\bm\gamma) = \bm{J}_i(\bm\gamma) \bm{N}_i(\bm\gamma)^T.
% \end{align}

Then, we can formulate corresponding hierarchically decoupled task-space velocities as follows:
\begin{align}
\bm{\nu}= 
		\begin{bmatrix}
			\bm{\nu}_1\\
			\bm{\nu}_2\\
			\bm{\nu}_3
		\end{bmatrix}
	= 
		\begin{bmatrix}
			\bm{\bar{J}}_1(\bm\gamma) \\
			\bm{\bar{J}}_2(\bm\gamma) \\
			\bm{\bar{J}}_3(\bm\gamma) 
		\end{bmatrix}
		\dot{\bm\gamma}
		=
	\bm{\bar{J}}(\bm\gamma)
	\dot{\bm\gamma}.
\end{align}
Here, $\bm{\nu}$ indicates the vector of the new local decoupled space velocities, and  $\bm{\bar{J}}(\bm\gamma)\in \mathbb{R}^{10 \times 10}$ is the extended Jacobian matrix that maps generalized velocities to the task velocities  $\bm{\nu}$. Note that $\bm{\bar{J}}(\bm\gamma)$ is invertible due to added 3rd task.

The new set of coordinates allows to receive hierarchically decoupled motion dynamics of the system (\ref{eq:dynamics_simplified_final3}):
%\small
\begin{align}
	\label{eq:dynamics_decoupled}
	\bm{\Lambda(\bm\gamma)}
	\bm{\dot\nu}
	+
	\bm{\mu(\bm\gamma,\dot{\bm\gamma}})
	\bm{\nu}
	= \bar{\bm{J}}(\bm\gamma)^{-T}(\bm{\tau}_\gamma-\bm{g_r(\bm\gamma))},
\end{align}
%\normalsize
where $\bm{\Lambda(\bm\gamma)} = (\bm{\bar{J}} \bM_r(\bq)^{-1}   \bm{\bar{J}}^T)^{-1} = diag(\bm{\Lambda}_1 ,\ \bm{\Lambda}_2 ,\ \bm{\Lambda}_3)$ contains decoupled task inertias, and %$\bm{\mu(\bm\gamma,\dot{\bm\gamma}}) = \bm{\Lambda}(\bm{\bar{J}} \bM'^{-1}\bC'  - \dot{\bm{\bar{J}}})\bm{\bar{J}}^{-1}$ is Coriolis-centrifugal matrix.
$\bm{\mu(\bm\gamma,\dot{\bm\gamma}}) =\bm{\bar{J}}^{-T}(\bC_r(\bm{\gamma}, \dot{\bm{\gamma}}) - \bM_r(\bq)\bm{\bar{J}}^{-1} \dot{\bar{\bm{J}}} )\bm{\bar{J}}^{-1}$ is transformed Coriolis-centrifugal matrix.

\subsection{Impedance whole-body control law formulation}
\label{subsec:wincharm_ctrl}

Since all tasks exploit the same $DOF$s, a hierarchical whole-body controller is applied in order to ensure that the main task is fulfilled without being disturbed by the second and third. The control law is formulated as follows:
\small
\begin{align}
		\label{eq:controllaw}
		\bm{\tau}_\gamma = 
		\begin{bmatrix}
			\btau_p\\
			\btau_m
		\end{bmatrix}
		=  \bm{g_r(\bm\gamma)} + \btau_{\mu} + \sum_{i=1}^{3}\bar{\bJ}^T_i\bF_{i,ctrl}%\sum_{i=2}^{3}\bm{N_i}\bm{J^T_i}\bm{F_{i,ctrl}}
		\end{align}	
\normalsize
%with	$ 
%\btau_{\mu} = \sum_{i=1}^{3} \left(\bar{\bJ}_i^T \left(\sum_{j=1}^{i-1}\bm{\mu}_{i,j} \bm{\nu}_j+\sum_{j=i+1}^{3}\bm{\mu}_{i,j} \bm{\nu}_j     \right)  \right).$
Here $\bF_{i,ctrl}$ is the control force (body wrench) for the $i-$th task before being projected in the null space, and $
\btau_{\mu} = \sum_{i=1}^{3} \left(\bar{\bJ}_i^T \left(\sum_{j=1}^{i-1}\bm{\mu}_{i,j} \bm{\nu}_j+\sum_{j=i+1}^{3}\bm{\mu}_{i,j} \bm{\nu}_j     \right)  \right)$ is applied in order to decouple task dynamics at the velocity level. It compensates for the off-blockdiagonal submatrices of the Coriolis-centrifugal matrix $\bm{\mu(\bm\gamma,\dot{\bm\gamma}})$: $\bm{\mu_{12}} \in \mathbb{R}^{6 \times 3}$, $\bm{\mu_{13}} \in \mathbb{R}^{6}$. %Thus, all task control forces $\bF_{i,ctrl}$ are expressed through all joint torques of the system in such a way that tasks are performed without interference according to hierarchy.
%This aspect establishes the foundation of the hierarchical whole-body control framework.

% By applying (\ref{eq:controllaw}) to the  (\ref{eq:dynamics_decoupled}), we receive fully decoupled dynamics equations as follows:
% \begin{align}
% 	\label{eq:dynamics_decoupled_fin_fin}
% 	\bm{\Lambda}_i(\bm\gamma)
% 	\bm{\dot\nu}_i
% 	+
% 	\bm{\mu}_{i,i}(\bm\gamma,\dot{\bm\gamma})
% 	\bm{\nu}_i
% 	= \bF_{i,ctrl} +\bar{\bm{J}}_i(\bm\gamma)^{-T}\bm{\tau}_{ext,{r_i}}
% \end{align}
% for $i \in \{1,\ 2 ,\ 3 \}$.

%With structural feasibility of all tasks and absence of external torques, the proposed control law (\ref{eq:controllaw}) guarantees the asymptotic stability of the equilibrium point for the main task and conditional stability for the second and third tasks as shown in \cite{dietrich2016whole}.

To accomplish the desired tasks and achieve compliant behaviour on all hierarchical levels, the Cartesian impedance control is exploited \cite{zhang2000}:
%\small
\begin{align}
	\label{eq:ctrl_forces_imp}
	\begin{aligned}
		\bF_{i,ctrl}
		=
		- \bm{K}_{P,i} \bm{p}_{i}
		- \bm{K}_{D,i}\bm{\dot{x}_{i}}.
	\end{aligned}
\end{align}
%\normalsize
Here, the matrices $\bm{K}_{(.)}$ are positive definite gain matrices (subscripts $P$, $D$ stand for positional stiffness and damping, respectively), $\bm{p}_{(.)}$ is the pose error between desired and current pose, and $\bm{\dot{x}_{i}}$ is the velocity of the i-th task in (\ref{eq:task_space}). 

%\SetTracking[spacing={-40*,0*,50*}]{encoding=OT1}{-23}
\SetTracking[spacing={}]{encoding=*}{-24}
\textls{
Formulated control law (\ref{eq:controllaw}) can be applied directly to our system for execution of all control tasks. However, the winch servos are position-controlled, so additional transformation is required.}%However, \red{while the robotic arm is torque controlled,} the winch servos are position-controlled. Therefore, additional transformation is required.}

\subsection{Admittance interface}
\label{subsec:adm}
Each winch takes the desired length of cable as an input signal. In order to operate the whole system at the torque level, the admittance interface was adapted \cite{dietrich2016whole, iskandar2019employing}. It takes commanded forces in joints as input to a virtual system with desired dynamics and produces the platform displacements as output. Exploitation of such an interface implies utilising the high gain motion controller for the winch actuation which can perfectly realize desired admittance dynamics despite external and internal disturbances, so no high-computational forward kinematics for cable-length feedback \cite{merlet2015forward} from winch servos is utilized as shown in Fig. \ref{fig:winch_control_diagram}.

Passing $\btau_p$ through the admittance interface with desired dynamics, corresponding displacement,  $\bm{x}_{p}^{des}$, is defined:
\begin{align}
	\label{eq:adm}
	%\bm{M_{adm}}\ddot{\bm{x}}\bm{_{p}^{des}} +  \bm{D_{adm}}\dot{\bm{x}}\bm{_{p}^{des}}  + \bm{K_{adm}}\bm{{x}_{p}^{des}}=  \bm\tau_p,
	\bm{M}_{adm}\ddot{\bm{x}}_{p}^{des} +  \bm{D}_{adm}\dot{\bm{x}}_{p}^{des} =  \bm\tau_p,
\end{align}
where $\bm{M}_{adm},\ \bm{D}_{adm} \in \mathbb{R}^{3 \times 3}$  %and $\bm{K_{adm}} \in \mathbb{R}^{3 \times 3}$
are positive inertia and damping diagonal matrices describing desired system dynamics.

Assuming that for embedded high-gain motion controller $\bm{x}_{p} \approx \bm{x}_{p}^{des}$, the overall system dynamics (\ref{eq:dynamics_simplified_final3}) can be rewritten:
\small
\begin{align}
	\label{eq:coupling}
&	\begin{bmatrix}
		\bm{M}_{adm} & \bm{0} \\ 
		\bm{M}_{pm}(\bm\gamma) & \bm{M}_{m}(\bm\gamma) \\
	\end{bmatrix} 
	\begin{bmatrix}
		\ddot{\bm{x}}_{p} \\ 
		\ddot{\bq}_{m} \\
	\end{bmatrix} 
	\\+ 
	& \begin{bmatrix}
		\bm{D}_{adm} & \bm{0} \\
		\bm{C}_{pm}(\bm\gamma,\ \dot{\bm\gamma}) & \bm{C}_m (\bm\gamma,\ \dot{\bm\gamma})\\ 
	\end{bmatrix}
	\begin{bmatrix}
		\dot{\bm{x}}_{p} \\
		\dot{\bq}_{m}\\
	\end{bmatrix} 
	+ 
	\begin{bmatrix}
        \bm{0}\\
	\bm{g}_m (\bm\gamma)
	\end{bmatrix}
	=
	\begin{bmatrix}
		\btau_{p} \\
		\btau_m
	\end{bmatrix}. \notag
\end{align}
\normalsize
Here, $\bM_{pm}$ and $\bC_{pm}$ are the inertia and Coriolis couplings between platform and manipulator which are submatrices of $\bM_r(\bm\gamma)$ and $\bC_r(\bm\gamma,\ \dot{\bm\gamma})$. % In order to estimate this term the force-torque sensor at the end effector or tension force sensor in the rigging cables can be installed. 
Following the \cite{dietrich2016whole}, control law (\ref{eq:controllaw}) should be extended with an additional term to remove coupling effect between manipulator and platform in (\ref{eq:coupling}) due to admittance dynamics.

\renewcommand{\thefigure}{5}
\begin{figure}[t]
      \centering
      \includegraphics[trim=0 0 0 0,clip, width=0.78\linewidth]{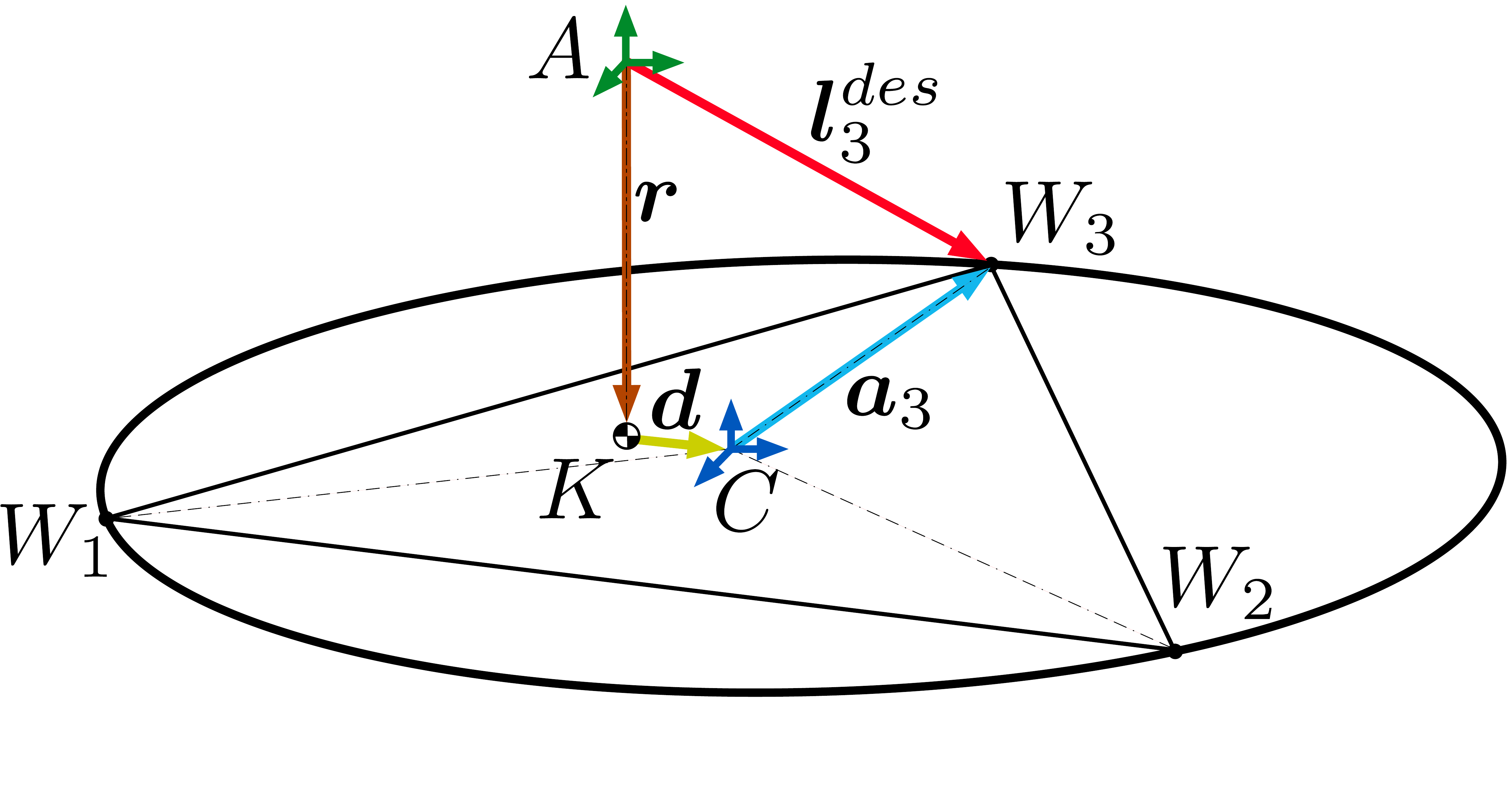}  
      \caption{Inverse kinematics.}
      \label{fig:ik}
\end{figure}

\renewcommand{\thefigure}{7}
\begin{figure*}[b]
      \centering
          \includegraphics[width=1\linewidth]{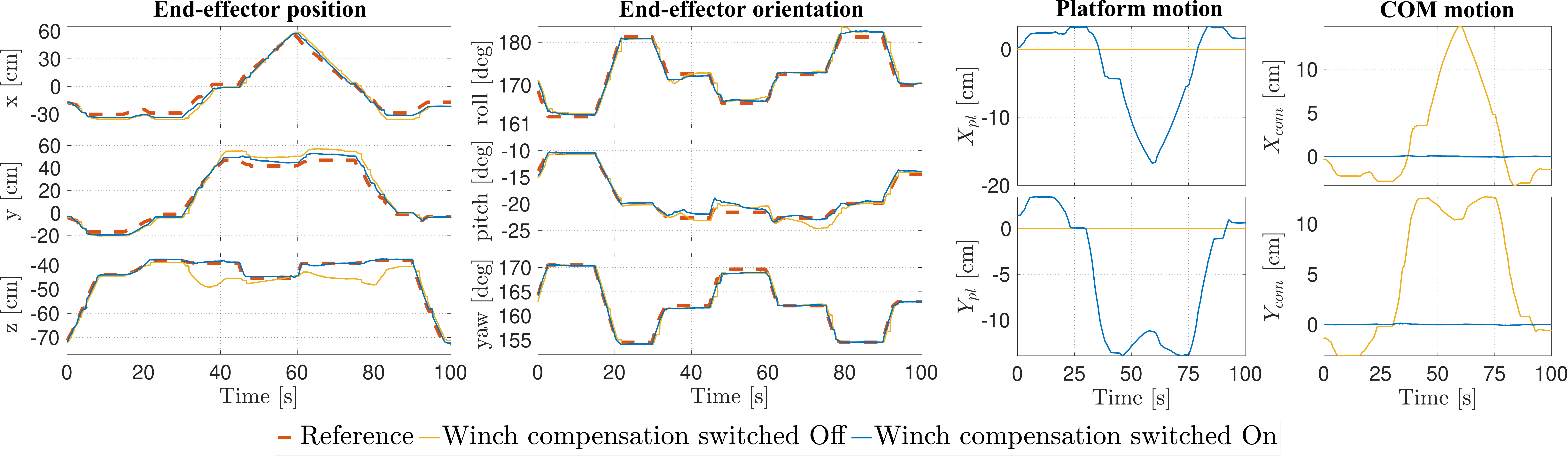} 
    \caption{Results of the first experiment: the end-effector follows the preprogrammed trajectory. The data is extracted from the model based on (\ref{eq:coupling}).}
    % change counter https://tex.stackexchange.com/questions/406918/custom-numbering-of-figures/406922
    \label{fig:first_exp}
\end{figure*}

% \renewcommand{\thefigure}{7}
% \begin{figure*}[b]
% 	\captionsetup[subfigure]{labelformat=empty}
% 	\centering
% 	\subcaptionbox{(a) First control task: end-effector pose  \label{fig:real_sys} }{\includegraphics[height=5.17cm,keepaspectratio]{images/1.pdf}}
% \subcaptionbox{(b)  Second task: platform and $COM$ motion  \label{fig:three_of_them} }{\includegraphics[height=5.17cm,keepaspectratio]{images/2.pdf}}
% \caption{Results of the first experiment: end-effector follows the preprogrammed trajectory. The data is extracted from the model.}
% 	\label{fig:first_exp}
% \end{figure*}
% \begin{align}
% 	\label{eq:fin_ctrl}
% 	\btau_{\gamma} =\bm{g_r(\bm\gamma)} + \btau_{\mu} + \btau_{comp} + \sum_{i=1}^{3}\bar{\bJ}^T_i\bF_{i,ctrl}.
% \end{align}

%\red{With structural feasibility of all tasks and absence of external torques, the proposed control law (\ref{eq:fin_ctrl}) guarantees the asymptotic stability of the equilibrium point for the main task and conditional stability for the second and third tasks maintaining the overall passivity, as shown in} \cite{dietrich2016whole, ott15}.

%It is worth noting that acceleration and velocities for $\btau_{comp}$ term can be estimated from (\ref{eq:adm}) without direct measurement.

As a final step, the calculated displacement, $\bm{x}^{des}_{p}$, should be transformed to the rigging cable lengths.

\subsection{Inverse kinematics}
\label{subsec:ik}
In order to control joints $\bm{x}_{p} = [x_c,\ y_c,\ z_c]^T$ via the lengths of the cables, the inverse kinematics for the rigging cable suspension is formulated. As shown in Fig. \ref{fig:ik}, % when pendulum joints are zeros,
the following vector loop can be constructed \cite{6907731}:
$
\bm{l}_i^{des}= \bm{r} + \bm{R}_p^w(\bm{d} + \bm{a}_i).
$
% \begin{align}
% 	\label{eq:ik}
% 	\bm{l}_i^{des}= \bm{r} + \bm{R}_p^w(\bm{d} + \bm{a}_i).
% \end{align}
Here, $\norm{\bm{l}^{des}_i}$ is the length of the $i-$cable. Constant vector $\bm{a}_i$ represents the location of the $i-$th cable start point, $W_i$, w.r.t. the platform geometric center, point $C$. Vector $\bm{r} = [0,\ 0,\ z_c]^T$ connects the suspension point $A$ with point $K$ which indicates the system $COM$, point $G$, projected at the platform plane, see Fig. \ref{fig:points}. Displacement between $K$ and $C$ is defined by vector $\bm{d}$. $\bm{R}_p^w$ is a rotation matrix representing orientation of the platform via roll and pitch angles. Since it is in our interests to keep these angles as zeros, we impose $\bm{R}_p^w=\bm{I}$ such that $\bm{R}_p^w\bm{d} =  [x_c,\ y_c,\ 0]^T$. Thus, presented $IK$ calculates such lengths which ensures horizontal $COM$ components under the suspension point while keeping horizontal orientation of the platform.

%Thus, for desired platform displacement $[x_c,\ y_c,\ z_c]^T$, we calculate only such cable lengths that guarantees zero angles. It is possible because inverse kinematics represents fully determined nonlinear system of equations  (\ref{eq:ik}), which can be solved for any desired quasi-state.

\renewcommand{\thefigure}{6}
\begin{figure}[t]
      \centering
      % include first image
      \includegraphics[width=0.95\linewidth]{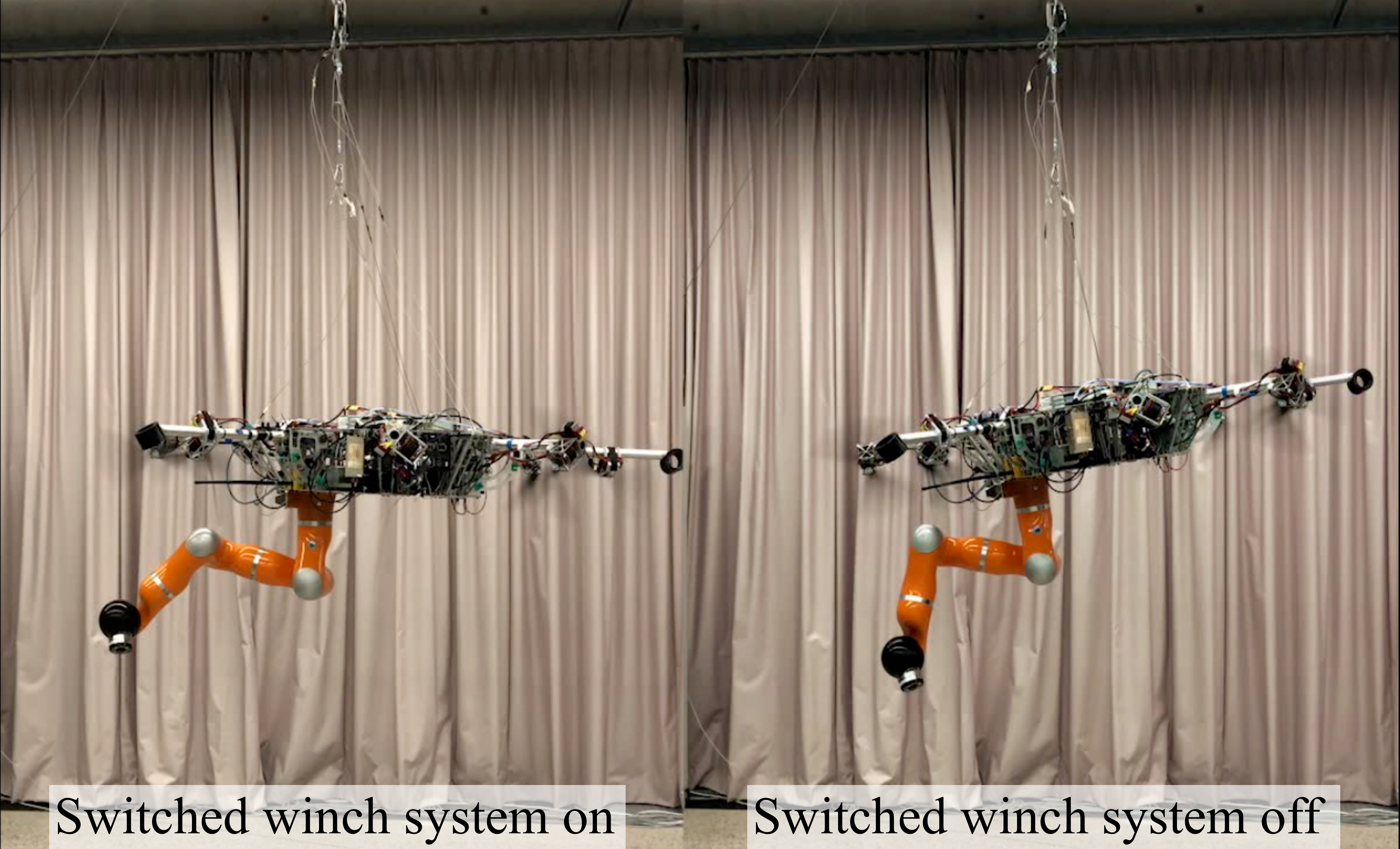}  
      \caption{The first experiment: the end-effector follows preprogrammed trajectory. The platform with switched winch system off tilts due to gravity.}
      \label{fig:platforms}
\end{figure}

\section{Experimental Validation}
\label{sec:val}
In order to investigate the winch system performance and validate the proposed control law (\ref{eq:controllaw}), we conducted two experiments. In both experiments, the platform was manually stabilized before the start, and propellers were used to keep the constant yaw angle without applying oscillation damping. The SAM dynamics (\ref{eq:coupling}) for reduced state $\bm{{\gamma}}$ was simulated onboard using the dynamics library based on \cite{garofalo2013closed}.%, \red{the platform was modeled as uniform cylinder}.

\subsection{Preprogrammed trajectory following}
\label{sec:hwbcperform}
During the first experiment, the manipulator end-effector moved along a preprogrammed trajectory, see Fig. \ref{fig:platforms}. Such a motion should force the system to tilt due to displaced $COM$ under the gravity effect. We conducted this experiment twice with switching the winch system on and off in order to compare the behavior. As a result, with switched winch-based actuation on, the end-effector reached desired position while $COM$ of the system remained under the suspension point keeping the platform horizontal. At the same time, the winch disabling led to the platform tilting.

Results of the first and second control tasks are shown in Fig. \ref{fig:first_exp}. Presented data is extracted directly from the real-time model based on (\ref{eq:coupling}). Since the model state does not take into account the tilting of the platform due to gravity, the real performance of the system with switched winches off for the first control task is even worse, see attached video for visual comparison.% \red{stress better that it is not a ground-truth, and that manipulation was autonomous, in real operation for the operator is a stress to operate on the platform that is tilting during time}

%As we can see, the proposed centralized whole-body controller with switched winch-based actuation on exposed a better performance for translational motion (Fig. \ref{fig:First_task1_xyz}) and comparable performance for the rotational motion of end-effector in terms of modeled joints, see Fig. \ref{fig:First_task1_rpy}. 
The second task is also fulfilled. With active winches, the platform kept horizontal orientation, while corresponding terms of the $COM$ had zero values. Compensation of the gravitational torque was performed by the platform motion. When we switched winches off and repeated the experiment, the result was the opposite.

\subsection{Pick and place of the cage}
\label{sec:pickplace}
In order to validate the applicability of the total framework, the pick and place task was conducted, see Fig. \ref{fig:pickandplace}. The task was to pick the empty cage from the metallic case and move it to the pipe. Such a cage might be utilised for transportation of mobile inspection robots. In this experiment, the end-effector was controlled remotely by an operator using a joystick with force feedback. It can be seen that during the whole mission, the platform kept its own orientation close to the horizontal. % while the height of the platform was adapting to assist the robotic arm: winches pulled the platform up and down in order to reduce the stretching of the manipulator to avoid singularity.% \red{manipulator workspace is restricted. It allows to avoid singular configurations for the robotic arm in some directions, in particular, fully stretched configuration in the vertical direction. }
The winch-based actuation fully provided designed capabilities, i.e., $COM$ in the horizontal plane was close to zero while the platform shifted in the horizontal plane in order to compensate for the disturbing gravitational torque, see Fig. \ref{fig:pickandplace_experiment_total}. It is worth noting that the platform precisely followed the command of admittance interface with selected gains: $\bM_{adm} = 0.8 \cdot eye(3)$, $\bm{D}_{adm}= 1.6 \cdot eye(3)$.
%\renewcommand{\thefigure}{14}
%       \includegraphics[width=1\linewidth]{images/exp_admittance.eps}  
% \caption{Tracking of the admittance interface commands \red{it is still can be seen in (b), so might be removed}Results of the third experiment at which the SAM picks the cage from the metallic box and places it on the pipe. Admittance and second, while first is clear since we manipulate using joystick not aunomously in constrast to the first experiment}
%     \label{fig:pickandplace_experiment_total}
% \end{figure}
\renewcommand{\thefigure}{8}
\begin{figure}[t]
      \centering
      % include first image
      \includegraphics[trim=0 1.116cm 0 0.346cm,clip,width=1\linewidth]{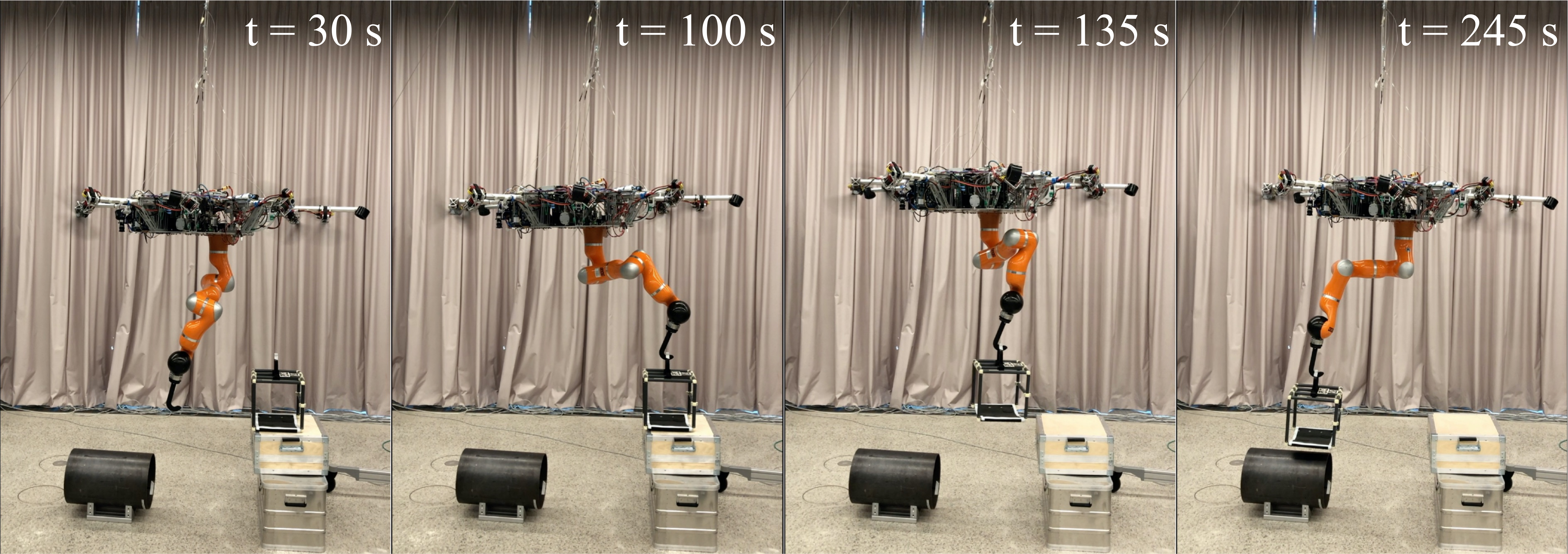}  
    \caption{The second experiment: deployment of the cage for mobile robot.}
    \label{fig:pickandplace}
\end{figure}
\renewcommand{\thefigure}{9}
\begin{figure}[t]
      \includegraphics[width=1\linewidth]{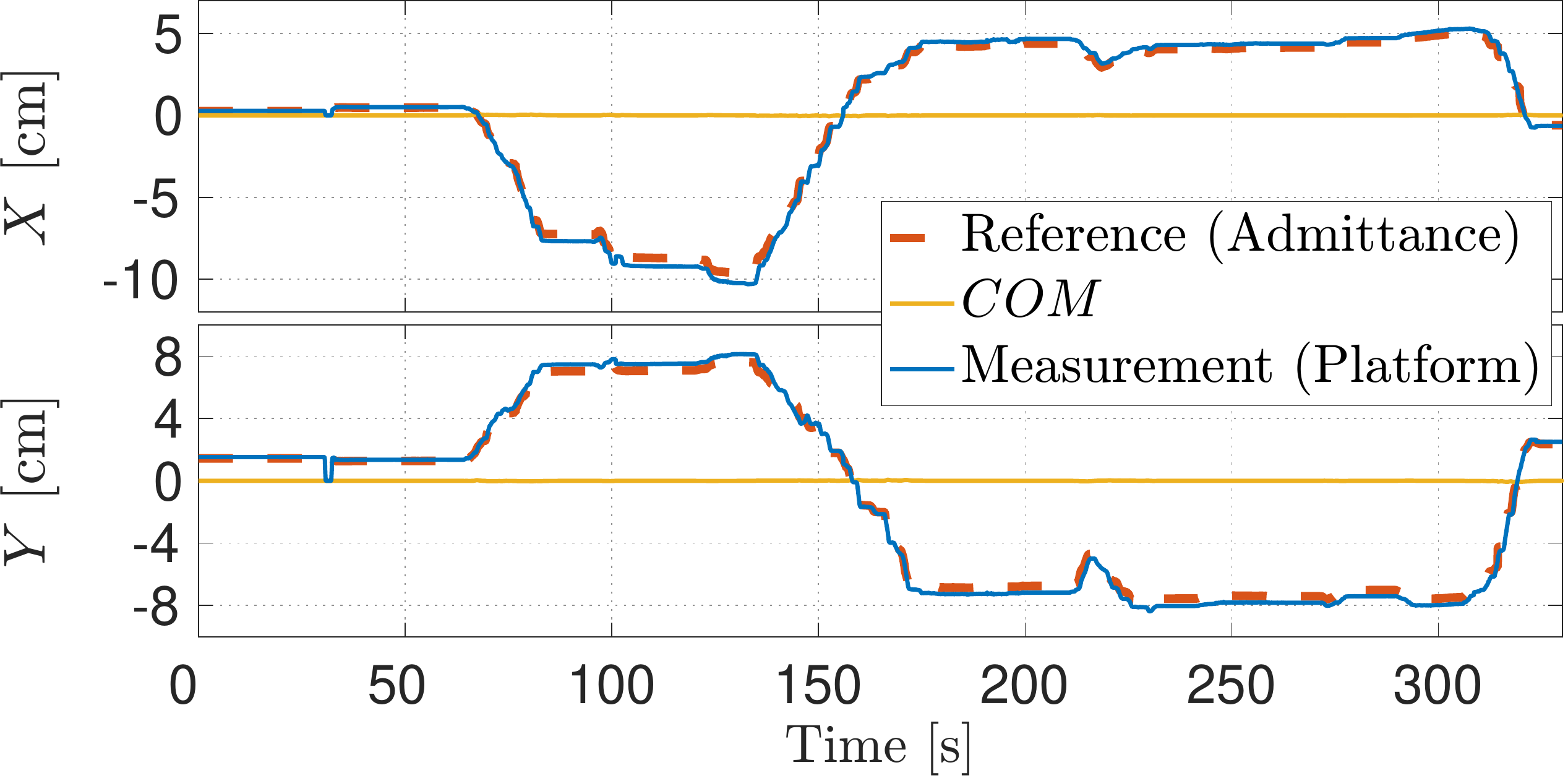}  
\caption{Results of the second experiment: pick and place of the cage. Platform follows admittance interface commands keeping zero horizontal terms of $COM$ ($\Delta x_{com} <$ 0.77 mm, $\Delta y_{com}<$ 0.85 mm, $RMSE_{x/y_{com}}$ = 0.13  mm). Platform displacement (measurement) was calculated in post processing using forward kinematics applied to the measured winch cable lengths.}
    \label{fig:pickandplace_experiment_total}
\end{figure}

\section{Conclusion}
\label{sec:end}

In this paper, we introduced the novel winch-based actuation for the cable-suspended aerial manipulator SAM. It helped to compensate for the disturbing gravitational torque by changing the length of suspension rigging cables. System closed-chain dynamics was transformed to equal serial-chain dynamics instantaneously feasible w.r.t. the defined holonomic constraints. It allowed to access the platform translational motion instead of the cable joints and to formulate two intuitive control tasks: to keep the desired pose of arm end-effector and to shift the system $COM$ under the suspension point. Both tasks were regulated by the impedance hierarchical whole-body controller with integrated admittance interface and inverse kinematics for the position-controlled winch servos. Received results demonstrated the applicability of the proposed actuation system as well as control framework to the real scenarios for implementing complex $AM$ missions. A further research could investigate the performance and energy cost optimization when both actuation systems are activated.% \red{energy cost has not be studied in the scope of the paper }
%%%%%%%%%%%%%%%%%%%%%%%%%%%%%%%%%%%%%%%%%%%%%%%%%%%%%%%%%%%%%%%%%%%%%%%%%%%%%%%%

%%%%%%%%%%%%%%%%%%%%%%%%%%%%%%%%%%%%%%%%%%%%%%%%%%%%%%%%%%%%%%%%%%%%18%%%%%%%%%%%%

%%%%%%%%%%%%%%%%%%%%%%%%%%%%%%%%%%%%%%%%%%%%%%%%%%%%%%%%%%%%%%%%%%%%%%%%%%%%%%%%

%%%%%%%%%%%%%%%%%%%%%%%%%%%%%%%%%%%%%%%%%%%%%%%%%%%%%%%%%%%%%%%%%%%%%%%%%%%%%%%%

\clearpage
\newpage

\bibliographystyle{IEEEtran.bst}
\bibliography{mybib.bib}

% Generated by IEEEtran.bst, version: 1.14 (2015/08/26)
\begin{thebibliography}{10}
\providecommand{\url}[1]{#1}
\csname url@samestyle\endcsname
\providecommand{\newblock}{\relax}
\providecommand{\bibinfo}[2]{#2}
\providecommand{\BIBentrySTDinterwordspacing}{\spaceskip=0pt\relax}
\providecommand{\BIBentryALTinterwordstretchfactor}{4}
\providecommand{\BIBentryALTinterwordspacing}{\spaceskip=\fontdimen2\font plus
\BIBentryALTinterwordstretchfactor\fontdimen3\font minus
  \fontdimen4\font\relax}
\providecommand{\BIBforeignlanguage}[2]{{%
\expandafter\ifx\csname l@#1\endcsname\relax
\typeout{** WARNING: IEEEtran.bst: No hyphenation pattern has been}%
\typeout{** loaded for the language `#1'. Using the pattern for}%
\typeout{** the default language instead.}%
\else
\language=\csname l@#1\endcsname
\fi
#2}}
\providecommand{\BIBdecl}{\relax}
\BIBdecl

\bibitem{ollero2021past}
A.~Ollero, M.~Tognon, A.~Suarez, D.~Lee, and A.~Franchi, ``Past, present, and
  future of aerial robotic manipulators,'' \emph{IEEE Transactions on
  Robotics}, vol.~38, no.~1, pp. 626--645, 2021.

\bibitem{ruggiero2018aerial}
F.~Ruggiero, V.~Lippiello, and A.~Ollero, ``Aerial manipulation: A literature
  review,'' \emph{Robotics and Automation Letters}, vol.~3, no.~3, pp.
  1957--1964, 2018.

\bibitem{khamseh2018aerial}
H.~B. Khamseh, F.~Janabi-Sharifi, and A.~Abdessameud, ``Aerial manipulation—a
  literature survey,'' \emph{Robotics and Autonomous Systems}, vol. 107, pp.
  221--235, 2018.

\bibitem{kudelina2020main}
K.~Kudelina, B.~Asad, T.~Vaimann, A.~Rassolkin, A.~Kallaste, and D.~V.
  Lukichev, ``Main faults and diagnostic possibilities of bldc motors,'' in
  \emph{2020 27th International Workshop on Electric Drives: MPEI Department of
  Electric Drives 90th Anniversary (IWED)}.\hskip 1em plus 0.5em minus
  0.4em\relax IEEE, 2020, pp. 1--6.

\bibitem{nath2017optimization}
J.~D. Nath and W.~K. Durfee, ``Optimization and design principles of a
  minimal-weight, wearable hydraulic power supply,'' in \emph{Dynamic Systems
  and Control Conference}, vol. 58271.\hskip 1em plus 0.5em minus 0.4em\relax
  American Society of Mechanical Engineers, 2017, p. V001T30A002.

\bibitem{haus2017}
T.~Haus, M.~Orsag, and S.~Bogdan, ``Mathematical modelling and control of an
  unmanned aerial vehicle with moving mass control concept,'' \emph{Journal of
  Intelligent \& Robotic Systems}, vol.~88, pp. 1--28, 03 2017.

\bibitem{kim2018oscillation}
M.~J. Kim, J.~Lin, K.~Kondak, D.~Lee, and C.~Ott, ``Oscillation damping control
  of pendulum-like manipulation platform using moving masses,''
  \emph{IFAC-PapersOnLine}, vol.~51, no.~22, pp. 465--470, 2018.

\bibitem{alakhras2022design}
A.~Al~Akhras, I.~H. Sattar, M.~Alvi, M.~W. Qanbar, M.~A. Jaradat, and
  M.~Alkaddour, ``The design of a lightweight cable aerial manipulator with a
  cog compensation mechanism for construction inspection purposes,''
  \emph{Applied Sciences}, vol.~12, no.~3, p. 1173, 2022.

\bibitem{pose2022adaptive}
C.~Pose, J.~Giribet, and I.~Mas, ``Adaptive center-of-mass relocation for
  aerial manipulator fault tolerance,'' \emph{IEEE Robotics and Automation
  Letters}, vol.~7, no.~2, pp. 5583--5590, 2022.

\bibitem{miyazaki2019long}
R.~Miyazaki, R.~Jiang, H.~Paul, Y.~Huang, and K.~Shimonomura, ``Long-reach
  aerial manipulation employing wire-suspended hand with swing-suppression
  device,'' \emph{IEEE Robotics and Automation Letters}, vol.~4, no.~3, pp.
  3045--3052, 2019.

\bibitem{Yiit2021NovelOA}
A.~Yiğit, M.~A. Perozo, L.~Cuvillon, S.~Durand, and J.~Gangloff, ``Novel
  omnidirectional aerial manipulator with elastic suspension: Dynamic control
  and experimental performance assessment,'' \emph{IEEE Robotics and Automation
  Letters}, vol.~6, pp. 612--619, 2021.

\bibitem{perozo2022optimal}
M.~A. Perozo, J.~Dussine, A.~Yi{\u{g}}it, L.~Cuvillon, S.~Durand, and
  J.~Gangloff, ``Optimal design and control of an aerial manipulator with
  elastic suspension using unidirectional thrusters,'' in \emph{2022
  International Conference on Robotics and Automation (ICRA)}.\hskip 1em plus
  0.5em minus 0.4em\relax IEEE, 2022, pp. 1976--1982.

\bibitem{sarkisov2019development}
Y.~S. Sarkisov, M.~J. Kim, D.~Bicego, D.~Tsetserukou, C.~Ott, A.~Franchi, and
  K.~Kondak, ``Development of sam: Cable-suspended aerial manipulator,'' in
  \emph{International Conference on Robotics and Automation (ICRA)}.\hskip 1em
  plus 0.5em minus 0.4em\relax IEEE, 2019, pp. 5323--5329.

\bibitem{6907731}
W.~{Kraus}, V.~{Schmidt}, P.~{Rajendra}, and A.~{Pott}, ``System identification
  and cable force control for a cable-driven parallel robot with industrial
  servo drives,'' in \emph{2014 IEEE International Conference on Robotics and
  Automation (ICRA)}, 2014, pp. 5921--5926.

\bibitem{1705581}
P.~{Bosscher}, A.~T. {Riechel}, and I.~{Ebert-Uphoff}, ``Wrench-feasible
  workspace generation for cable-driven robots,'' \emph{IEEE Transactions on
  Robotics}, vol.~22, no.~5, pp. 890--902, 2006.

\bibitem{begey2018dynamic}
J.~Begey, L.~Cuvillon, M.~Lesellier, M.~Gouttefarde, and J.~Gangloff, ``Dynamic
  control of parallel robots driven by flexible cables and actuated by
  position-controlled winches,'' \emph{IEEE Transactions on Robotics}, vol.~35,
  no.~1, pp. 286--293, 2018.

\bibitem{FREITAS2011}
G.~M. Freitas, A.~C. Leite, and F.~Lizarralde,
  ``\BIBforeignlanguage{en}{Kinematic control of constrained robotic
  systems},'' \emph{\BIBforeignlanguage{en}{Sba: Controle \& Automacao
  Sociedade Brasileira de Automatica}}, vol.~22, pp. 559 -- 572, 12 2011.

\bibitem{mcgrath2017lagrange}
M.~McGrath, D.~Howard, and R.~Baker, ``A lagrange-based generalised formulation
  for the equations of motion of simple walking models,'' \emph{Journal of
  biomechanics}, vol.~55, pp. 139--143, 2017.

\bibitem{murray1989dynamic}
J.~J. Murray and G.~H. Lovell, ``Dynamic modeling of closed-chain robotic
  manipulators and implications for trajectory control,'' \emph{IEEE
  Transactions on Robotics and Automation}, vol.~5, no.~4, pp. 522--528, 1989.

\bibitem{coelho2021hierarchical}
A.~Coelho, Y.~S. Sarkisov, J.~Lee, R.~Balachandran, A.~Franchi, K.~Kondak, and
  C.~Ott, ``Hierarchical control of redundant aerial manipulators with enhanced
  field of view,'' in \emph{2021 International Conference on Unmanned Aircraft
  Systems (ICUAS)}.\hskip 1em plus 0.5em minus 0.4em\relax IEEE, 2021, pp.
  994--1002.

\bibitem{sarkisov20}
Y.~Sarkisov, M.~J. Kim, A.~Coelho, D.~Tsetserukou, C.~Ott, and K.~Kondak,
  ``Optimal oscillation damping control of cable-suspended aerial manipulator
  with a single imu sensor,'' in \emph{2020 IEEE International Conference on
  Robotics and Automation (ICRA)}, 2020.

\bibitem{ibrahim2007kinematic}
O.~Ibrahim and W.~Khalil, ``Kinematic and dynamic modeling of the 3-rps
  parallel manipulator,'' in \emph{Proceedings of the 12th IFToMM World
  congress, France}, 2007, pp. 18--21.

\bibitem{my2019kinematic}
C.~A. My and V.~M. Hoan, ``Kinematic and dynamic analysis of a serial
  manipulator with local closed loop mechanisms,'' \emph{Vietnam Journal of
  Mechanics}, vol.~41, no.~2, pp. 141--155, 2019.

\bibitem{zambella2019dynamic}
G.~Zambella, G.~Lentini, M.~Garabini, G.~Grioli, M.~G. Catalano, A.~Palleschi,
  L.~Pallottino, A.~Bicchi, A.~Settimi, and D.~Caporale, ``Dynamic whole-body
  control of unstable wheeled humanoid robots,'' \emph{IEEE Robotics and
  Automation Letters}, vol.~4, no.~4, pp. 3489--3496, 2019.

\bibitem{dietrich2019hierarchical}
A.~Dietrich and C.~Ott, ``Hierarchical impedance-based tracking control of
  kinematically redundant robots,'' \emph{IEEE Transactions on Robotics},
  vol.~36, no.~1, pp. 204--221, 2019.

\bibitem{ott15}
C.~Ott, A.~Dietrich, and A.~Albu-Sch{\"a}ffer, ``Prioritized multi-task
  compliance control of redundant manipulators,'' \emph{Automatica}, vol.~53,
  pp. 416--423, 2015.

\bibitem{dietrich2016whole}
A.~Dietrich, K.~Bussmann, F.~Petit, P.~Kotyczka, C.~Ott, B.~Lohmann, and
  A.~Albu-Sch{\"a}ffer, ``Whole-body impedance control of wheeled mobile
  manipulators,'' \emph{Autonomous Robots}, vol.~40, no.~3, pp. 505--517, 2016.

\bibitem{lee20}
J.~Lee, R.~Balachandran, Y.~Sarkisov, M.~De~Stefano, A.~Coelho, K.~Shinde,
  M.~J. Kim, R.~Triebel, and K.~Kondak, ``Visual-inertial telepresence for
  aerial manipulation,'' in \emph{2020 IEEE International Conference on
  Robotics and Automation (ICRA)}, 2020.

\bibitem{dietrich2013multi}
A.~Dietrich, C.~Ott, and A.~Albu-Sch{\"a}ffer, ``Multi-objective compliance
  control of redundant manipulators: Hierarchy, control, and stability,'' in
  \emph{2013 IEEE/RSJ International Conference on Intelligent Robots and
  Systems}.\hskip 1em plus 0.5em minus 0.4em\relax IEEE, 2013, pp. 3043--3050.

\bibitem{khatib1987unified}
O.~Khatib, ``A unified approach for motion and force control of robot
  manipulators: The operational space formulation,'' \emph{IEEE Journal on
  Robotics and Automation}, vol.~3, no.~1, pp. 43--53, 1987.

\bibitem{zhang2000}
S.~Zhang and E.~D. Fasse, ``Spatial compliance modeling using a
  quaternion-based potential function method,'' \emph{Multibody System
  Dynamics}, vol.~4, no.~1, pp. 75--101, 2000.

\bibitem{iskandar2019employing}
M.~Iskandar, G.~Quere, A.~Hagengruber, A.~Dietrich, and J.~Vogel, ``Employing
  whole-body control in assistive robotics,'' in \emph{IEEE International
  Conference on Intelligent Robots and Systems}, 2019, pp. 5643--5650.

\bibitem{merlet2015forward}
J.-P. Merlet and J.~Alexandre-dit Sandretto, ``The forward kinematics of
  cable-driven parallel robots with sagging cables,'' in \emph{Cable-Driven
  Parallel Robots}.\hskip 1em plus 0.5em minus 0.4em\relax Springer, 2015, pp.
  3--15.

\bibitem{garofalo2013closed}
G.~Garofalo, C.~Ott, and A.~Albu-Schaffer, ``On the closed form computation of
  the dynamic matrices and their differentiations,'' in \emph{IEEE/RSJ
  International Conference on Intelligent Robots and Systems (IROS)}, 2013, pp.
  2364--2359.

\end{thebibliography}

\end{document}